\documentclass[journal]{IEEEtran}
\usepackage{epsfig}
\usepackage{multirow}
\usepackage{stackengine}
\usepackage{footnote}
\usepackage{url}
\newcommand{\urlwofont}[1]{\urlstyle{same}\url{#1}}
\usepackage{cite}

\ifCLASSINFOpdf
\else
\fi
\usepackage{amsmath}
  \usepackage[caption=false,font=footnotesize]{subfig}
\begin{document}
%
\title{Adaptive Detrending to Accelerate Convolutional Gated Recurrent Unit Training for Contextual Video Recognition}
%
%
%

\author{Minju Jung,
        Haanvid Lee,
        and Jun Tani
\thanks{This work was supported by the National Research Foundation of Korea (NRF) grant funded by the Korea government (MSIP) (No. 2014R1A2A2A01005491).}
\thanks{M. Jung and H. Lee are with the Department
of Electrical Engineering, Korea Advanced Institute of Science and Technology, Daejeon, Korea (e-mail: minju5436@gmail.com; haanvidlee@gmail.com).}
\thanks{J. Tani is with the Department
of Electrical Engineering, Korea Advanced Institute of Science and Technology, Daejeon, Korea, and also with the Cognitive Neurorobotics Research Unit, Okinawa Institute of Science and Technology Graduate University, Okinawa, Japan (e-mail: tani1216jp@gmail.com). Correspondence should be sent to J. Tani.}
}

\maketitle

\begin{abstract}
Based on the progress of image recognition, video recognition has been extensively studied recently. However, most of the existing methods are focused on short-term but not long-term video recognition, called contextual video recognition. To address contextual video recognition, we use convolutional recurrent neural networks (ConvRNNs) having a rich spatio-temporal information processing capability, but ConvRNNs requires extensive computation that slows down training.
In this paper, inspired by the normalization and detrending methods, we propose \textit{adaptive detrending} (AD) for temporal normalization in order to accelerate the training of ConvRNNs, especially for convolutional gated recurrent unit (ConvGRU). AD removes internal covariate shift within a sequence of each neuron in recurrent neural networks (RNNs) by subtracting a trend.
In the experiments for contextual recognition on ConvGRU, the results show that (1) ConvGRU clearly outperforms the feed-forward neural networks, (2) AD consistently offers a significant training acceleration and generalization improvement, and (3) AD is further improved by collaborating with the existing normalization methods. 
\end{abstract}


%
\IEEEpeerreviewmaketitle

\section{Introduction}
%
%
%
%

\begin{figure}[t]
\begin{center}
\includegraphics[width=3in]{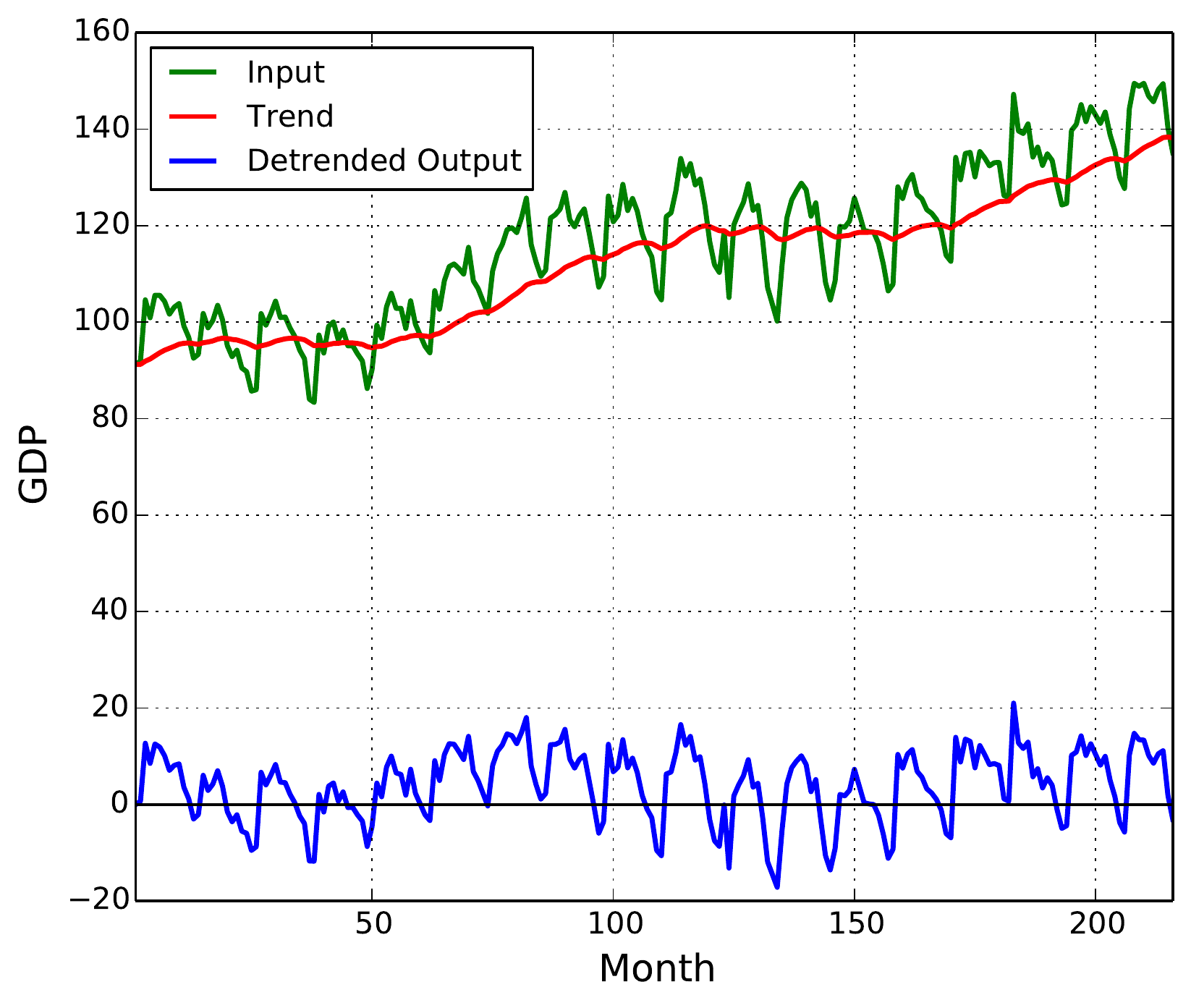}
\caption[Caption title in LOF]{Example of conventional detrending with Brazilian GDP. The detrended output is obtained by subtracting the trend from the original input. In this example, we use an exponential moving average (EMA) with a fixed decay factor of 0.95 to define the trend.}
\label{fig:Detrend}
\end{center} 
\end{figure}

\IEEEPARstart{C}{onvolutional} neural networks (CNNs) \cite{LeCun:1988} show remarkable performance on the ImageNet challenge dataset having 1.2 million training images and 1000 classes \cite{Krizhevsky:2012}. Encouraged by the success of CNNs, there are several approaches try to employ the spatial information processing capability of CNNs in video recognition tasks\cite{Simonyan:2014,Tran:2015}. Among the networks extending CNNs for video recognition, two-stream CNNs \cite{Simonyan:2014} and convolutional 3D (C3D) \cite{Tran:2015} are the most popularly used networks.
Two-stream CNNs are composed of a spatial-stream network receiving single RGB frame and a temporal-stream network receiving stacked optical flow over several frames and combine the classification scores from spatial- and temporal-stream networks, respectively.
C3D extends 2D convolution to 3D convolution by adding the time as a third dimension and receives stacked consecutive RGB frames.
However, both networks use a stacking strategy that utilizes only limited temporal correlations within stacked frames to recognize videos. Once the temporal window is slid to a next position, the processed information of the previous stack is completely dropped from a network, and it creates the problem of contextual recognition, which requires the extraction of long-range temporal correlations as mentioned by Jung et al. \cite{Jung:2015}.

Convolutional recurrent neural networks (ConvRNNs) have recently been introduced to merge the spatial and temporal information processing capabilities of CNNs and recurrent neural networks (RNNs), respectively, by replacing weight multiplication of RNNs with convolution \cite{Shi:2015,Ballas:2015,Kalchbrenner:2016}. By extracting spatio-temporal features hierarchically, ConvRNNs are beneficial to handle complex problems in the space-time domain, such as precipitation nowcasting \cite{Shi:2015}, video recognition \cite{Ballas:2015}, and video prediction \cite{Kalchbrenner:2016}.
Also, problems only for the spatial domain can be dealt with by ConvRNNs in an iterative manner \cite{Romera-Paredes:2016}. For example, in instance segmentation, ConvRNNs sequentially segment one instance of an image at a time \cite{Romera-Paredes:2016}.
In this paper, we use ConvRNNs for contextual video recognition. 
However, although ConvRNNs are a useful tool, the training of ConvRNNs is painfully slower than the feed-forward CNNs that receive a single frame or stacked multiple frames for video recognition because the additional computation of recurrent connections is required, and it is hard to parallelize the ConvRNNs computation across time. Therefore, finding the way to achieve a faster learning convergence is crucial for ConvRNNs.

Loffe and Szegedy \cite{Ioffe:2015} argue that internal covariate shift induces the degradation of training speed in feed-forward neural networks including multi-layer perceptron (MLP) and CNNs, and propose batch normalization (BN) that normalizes the input distribution of a neuron over a mini-batch. BN successfully removes internal covariate shift, so that BN significantly accelerates the training with improved generalization in feed-forward neural networks. Hence, BN has become a standard for the training of feed-forward neural networks. There are a few studies to bring the power of BN into RNNs because unrolled RNNs over time can be seen as deep neural networks in terms of time as well as depth \cite{Laurent:2015,Cooijmans:2016}. However, BN is not fully fitted with RNNs regardless of computing global statistics along the time domain \cite{Laurent:2015} or local statistics at each time step \cite{Cooijmans:2016}. Using global statistics ignores different statistics at each time step, and using local statistics is insufficient to properly handle training sequences having different lengths in a mini-batch. To eliminate the dependencies between samples in a mini-batch that make BN hard to apply to RNNs, layer normalization (LN) \cite{Ba:2016} computes statistics over all neurons in each layer and accelerates the training of RNNs and MLP, but not for CNNs. However, both BN limited to RNNs and LN limited to CNNs are not generally applied to ConvRNNs.

To find the method to accelerate the training of ConvRNNs, we focus on the time domain, which is not dealt with in the above studies.
Detreding is a method to transform non-stationary time series to stationary ones by taking out a trend because a lot of time series analysis and forecasting methods only can be applied to stationary time series. See the example of detrending with Brazilian GDP\footnote{\urlwofont{http://www2.stat.duke.edu/~mw/data-sets/ts_data/brazil_econ}} in Fig. \ref{fig:Detrend}.
Inspired by the concept of detrending, we apply the detrending method to normalize the sequences of neurons in RNNs. We notice that the hidden state of gated recurrent unit (GRU) \cite{Cho:2014} can be considered as a trend, which can be approximated by the form of an exponential moving average with an adaptively changing decay factor. Based on this key finding, we propose a novel temporal normalization method, called \textit{adaptive detrending} (AD), for convolutional gated recurrent unit (ConvGRU), which is a variant of ConvRNNs extended from GRU. The contributions of AD are fourfold as follows:
\begin{itemize}
  \item AD is easy to implement and has cheap computational cost and less memory consumption.
  \item AD eliminates internal covariate shift that has occurred through time.
  \item Because of the decay factor adaptability, AD controls the degree of detrending (or normalization).
  \item In ConvGRU, AD shows strong synergy with the existing normalization methods.
\end{itemize}

\section{Background}
\label{sec:background}
\subsection{Batch Normalization}
Internal covariate shift has been known that slows down the training of deep neural networks because the distribution of each layer's inputs is continuously changed as the parameters of the lower layers are updated during training. Batch normalization (BN) \cite{Ioffe:2015} is recently proposed to reduce internal covariate shift by normalizing network activations as follows: 
\begin{equation}
\label{BN:mean}
    {\mu} = \frac{1}{m} \sum_{i=1}^{m} {x_{i}}
\end{equation}
\begin{equation}
\label{BN:varaince}
    {\sigma}^{2} = \frac{1}{m} \sum_{i=1}^{m} {{(x_{i} - \mu)}^{2}}
\end{equation}
\begin{equation}
\label{BN:normalization}
    {\hat{x}_{i}} = \frac{x_{i} - \mu}{\sqrt{{\sigma}^{2} + \epsilon} }
\end{equation}
\begin{equation}
\label{BN:affine}
    {{y}_{i}} = \gamma \hat{x}_{i} + \beta
\end{equation}
where $x$ is the activations of a neuron in a mini-batch of size $m$, $\mu$ and $\sigma^2$ is the mean and variance of a mini-batch, $\hat{x}$ is the normalized inputs, $\epsilon$ is an infinitesimal constant for numerical stability, and $y$ is an affine transformation of the normalized inputs $\hat{x}$. 
The input distribution to a layer is transformed to the fixed distribution with a zero mean and unit variance, regardless of the change in parameters of all layers below during training. Also, an affine transformation with two learnable parameters $\gamma$ and $\beta$ follows the normalization to recover the original activations when it is required. BN has been shown to accelerate the training and improve the generalization of CNNs on ImageNet classification.



Inspired by success of BN in feed-forward neural networks, BN has been applied to recurrent neural networks (RNNs) to get training speed-up and generalization \cite{Laurent:2015,Cooijmans:2016}. One applies BN only on the vertical (or input-to-hidden) connection not on the horizontal (or hidden-to-hidden) connection because the repeated rescaling for the horizontal connection induces gradient vanishing and exploding problems \cite{Laurent:2015}. Also, the mean and variance for BN are computed by averaging along not only the mini-batch axis but also the time axis, which is called sequence-wise normalization. On the other hand, Cooijmans et al. \cite{Cooijmans:2016} show that (1) BN applied on not only the vertical connection but also the horizontal connection is highly beneficial, and (2) using statistics for each time step separately is good to preserve information of the initial transient phase, which is called step-wise normalization. However, estimation of statistics at each time step becomes poorer along the time axis because a mini-batch configuration for training sequences having different lengths involves zero padding to fill up after shorter sequences are finished. In addition, statistics for each time step are estimated only until the length of the longest training sequence $T_{max}$ is reached. During test phase, longer test sequences than the longest training sequence cannot utilize accurate statistics for normalization beyond $T_{max}$.

\subsection{Layer Normalization}
\label{sec:LN}
To overcome the limitations of BN when it is applied to RNNs, Ba et al. \cite{Ba:2016} introduce a new method called layer normalization (LN). LN has the same form as that of Cooijmans et al. \cite{Cooijmans:2016} except that LN normalizes over the spatial axis, rather than the mini-batch axis. The assumption of LN is that the changing of the layer's outputs will be highly correlated with the changing of the next layer's summed inputs. Hence, LN takes all activations in each layer on a single training data to estimate statistics. By estimating statistics over a layer not a mini-batch, LN can properly estimate statistics at each time step regardless of sequence length variability in a mini-batch. In the experiments with RNNs, LN shows a faster convergence and better generalization than the baseline and other normalization methods, especially for long sequences and small mini-batches. 

However, it does not work well when LN is used for CNNs. The authors report that LN is better than the baseline without normalization, but not for BN. They hypothesize that neurons in a layer have different statistics caused by spatial topology in a feature map, so that the assumption of LN cannot be supported in CNNs. We agree that all neurons from a layer are normalized with the same statistics is not the best way for CNNs, but the reason might be different statistics over feature maps, not within a feature map because BN successfully works for CNNs by estimating statistics of each feature map.

\section{Model}
\subsection{Gated Recurrent Unit}
Standard recurrent neural networks (RNNs) are extended from feed-forward networks by adding a recurrent connection to handle sequential data. RNNs consist of three layers: the input layer $\mathbf{x}$, hidden layer $\mathbf{h}$, and output layer $\mathbf{y}$. RNNs are able to handle sequential data because the hidden layer receives both the current input from the input layer and its own previous state through a recurrent connection as follows:
\begin{equation}
\label{RNN:hidden}
    \mathbf{h}_{t} = g(\mathbf{W}_{h}\mathbf{x}_{t} + \mathbf{U}_{h}\mathbf{h}_{t-1} + \mathbf{b}_{h})
\end{equation}
\begin{equation}
    \mathbf{y}_{t} = f(\mathbf{W}_{y}\mathbf{h}_{t} + \mathbf{b}_{y})
\end{equation}
where $g(\cdot)$ and $f(\cdot)$ are element-wise non-linear activation functions for the hidden and output layers, respectively, and $\mathbf{W}$, $\mathbf{U}$, and $\mathbf{b}$ represent the learnable parameters of RNNs as follows: forward connection weights, recurrent connection weights, and biases, respectively.

However, standard RNNs are hard to capture long-term dependencies because of the gradient vanishing and exploding problems \cite{Hochreiter:1991,Bengio:1994}. Gated recurrent unit (GRU) is proposed by Cho et al. \cite{Cho:2014} to overcome the gradient vanishing problem. It shares the same gating mechanism of long short-term memory (LSTM) \cite{Hochreiter:1997}, but has a more simple architecture by eliminating the output gate and modifying some other parts in LSTM. Specifically, GRU has two gating units, called a reset gate $\mathbf{r}$ and an update gate $\mathbf{z}$. The hidden state $\mathbf{h}_{t}$ at each time step $t$ is calculated by the form of leaky integrator with adaptive time constant determined by the update gate $\mathbf{z}$. In other words, the hidden state $\mathbf{h}_{t}$ is a linear interpolation between the previous hidden state $\mathbf{h}_{t-1}$ and the candidate hidden state $\tilde{\textbf{h}}_{t}$ weighted by the update gate $\mathbf{z}$ defined as follows:
\begin{equation}
\label{GRU:hidden}
    \mathbf{h}_{t} = \mathbf{z}_{t} \odot \tilde{\textbf{h}}_{t} + (1-\mathbf{z}_{t}) \odot \mathbf{h}_{t-1} 
\end{equation}
\begin{equation}
\label{GRU:updateGate}
    \mathbf{z}_{t} = \sigma(\mathbf{W}_{z}\mathbf{x}_{t} + \mathbf{U}_{z}\mathbf{h}_{t-1} + \mathbf{b}_{z})
\end{equation}
where $\sigma(\cdot)$ is a sigmoid function and $\odot$ is an element-wise multiplication.

The candidate hidden state $\tilde{\textbf{h}}_{t}$ at each time step $t$ is calculated similarly to that of the hidden layer in standard RNNs as shown in (\ref{RNN:hidden}). However, unlike standard RNNs, the reset gate $\mathbf{r}$ determines how much the previous hidden state $\mathbf{h}_{t-1}$ affects the candidate hidden state $\tilde{\textbf{h}}_{t}$ as follows:
\begin{equation}
\label{GRU:newHidden}
    \tilde{\textbf{h}}_{t} = \tanh(\mathbf{W}_{h}\mathbf{x}_{t} + \mathbf{r}_{t} \odot \mathbf{U}_{h}\mathbf{h}_{t-1} + \mathbf{b}_{h})
\end{equation}
\begin{equation}
\label{GRU:resetGate}
    \mathbf{r}_{t} = \sigma(\mathbf{W}_{r}\mathbf{x}_{t} + \mathbf{U}_{r}\mathbf{h}_{t-1} + \mathbf{b}_{h}) 
\end{equation}

\subsection{Gated Recurrent Unit Normalization in the Spatial Domain}
\label{sec:norm}
In this paper, following Ba et al. \cite{Ba:2016}, we apply recurrent batch normalization (recurrent BN) \cite{Cooijmans:2016} and layer normalization (LN) \cite{Ba:2016} to GRU.
We refer to recurrent BN and LN as spatial normalization methods to differentiate from the proposed normalization method in the time domain. The following equations represent GRU normalization in the spatial domain:
\begin{equation}
\label{Norm:resetGate}
    \mathbf{r}_{t} = \sigma(\textrm{N}_{\boldsymbol\gamma,\boldsymbol\beta}(\mathbf{W}_{r}\mathbf{x}_{t}) + \textrm{N}_{\boldsymbol\gamma}(\mathbf{U}_{r}\mathbf{h}_{t-1})) 
\end{equation}
\begin{equation}
\label{Norm:updateGate}
    \mathbf{z}_{t} = \sigma(\textrm{N}_{\boldsymbol\gamma,\boldsymbol\beta}(\mathbf{W}_{z}\mathbf{x}_{t}) + \textrm{N}_{\boldsymbol\gamma}(\mathbf{U}_{z}\mathbf{h}_{t-1}))
\end{equation}
\begin{equation}
\label{Norm:newHidden}
    \tilde{\textbf{h}}_{t} = \tanh(\textrm{N}_{\boldsymbol\gamma,\boldsymbol\beta}(\mathbf{W}_{h}\mathbf{x}_{t}) + \mathbf{r}_{t} \odot \textrm{N}_{\boldsymbol\gamma}(\mathbf{U}_{h}\mathbf{h}_{t-1}))
\end{equation}
\begin{equation}
\label{Norm:hidden}
    \mathbf{h}_{t} = \mathbf{z}_{t} \odot \tilde{\textbf{h}}_{t} + (1-\mathbf{z}_{t}) \odot \mathbf{h}_{t-1} 
\end{equation}
where $\textrm{N}_{\boldsymbol\gamma,\boldsymbol\beta}(\cdot)$ represents the normalization followed by an affine transformation with two learnable parameters (gain $\boldsymbol\gamma$ and bias $\boldsymbol\beta$) for recurrent BN and LN, and $\textrm{N}_{\boldsymbol\gamma}(\cdot)$ is the same as $\textrm{N}_{\boldsymbol\gamma,\boldsymbol\beta}(\cdot)$ except for an affine transformation with only the gain $\boldsymbol\gamma$ to remove the bias redundancy within an equation. For the same reason as $\textrm{N}_{\boldsymbol\gamma}(\cdot)$, the biases of the original GRU equations are removed. 

\subsection{Adaptive Detrending}

\begin{figure}[t]
\centering
\includegraphics[width=\linewidth]{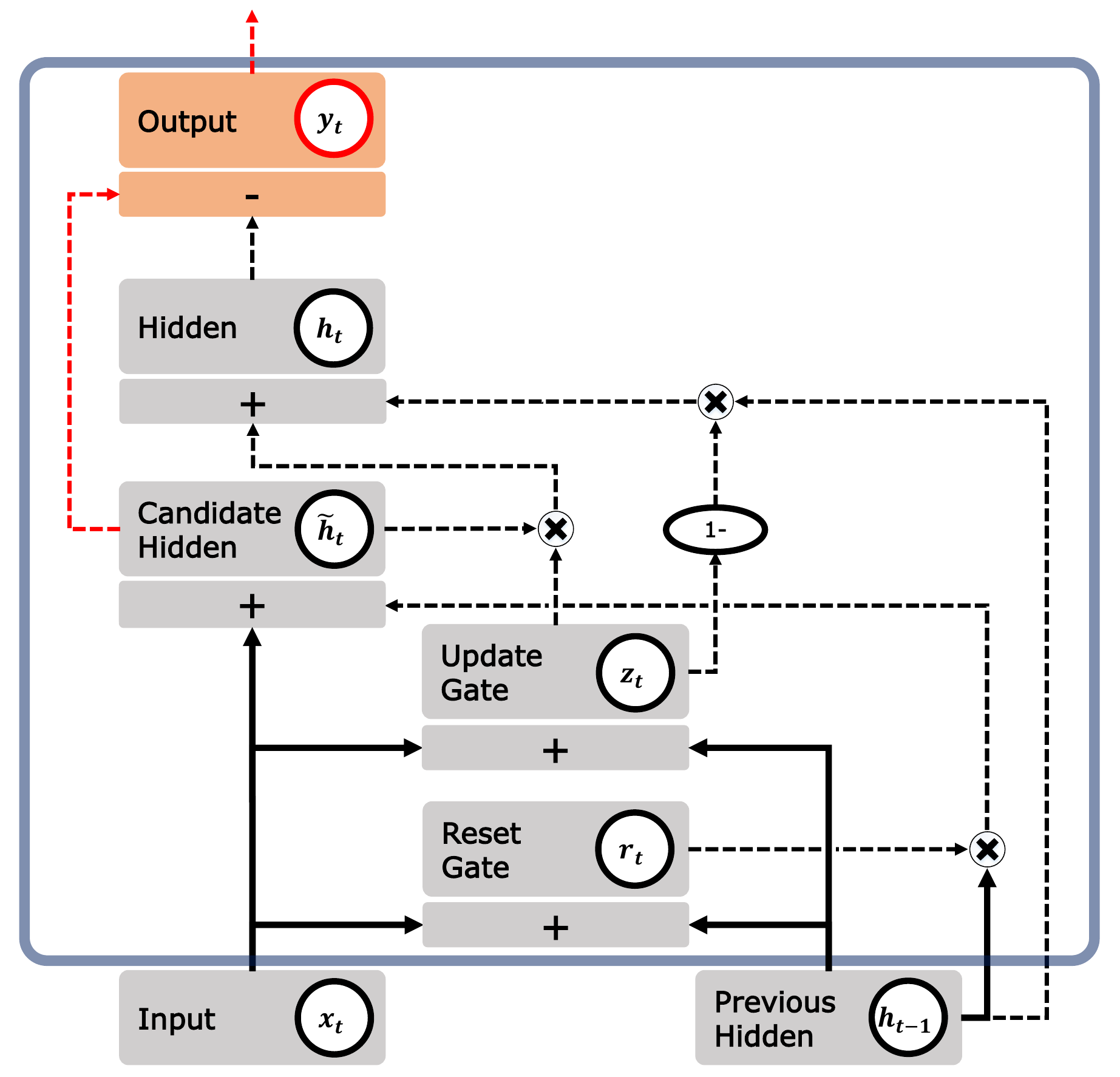}
\caption{Schematic of gated recurrent unit (GRU) with adaptive detrending (AD). The gray modules and black lines are for GRU, and the light red modules and red lines are newly added for AD. Solid lines represent weight multiplication operations and dashed lines represent element-wise operations.}
\label{fig:Model}
\end{figure}

The spatial normalization methods have achieved the training acceleration and performance improvement on complex sequential tasks, such as language modeling, by normalizing in a step-wise manner \cite{Cooijmans:2016,Ba:2016}.
However, the spatial normalization methods have a limitation. Although statistics are estimated at each time separately in order to capture the initial transient, each statistics estimation is only based on the current neural activations. It is not the best way for RNNs because the true statistics of RNNs at the current step inherently depends on those of previous steps. Hence, statistics estimation for RNNs should follow how RNNs generate statistics over time. Specifically, currently estimated statistics need to participate in the next estimation. The condition of statistics estimation for RNNs reminds us of a moving average (MA). 



In statistics, MA is widely used to filter out noise or fluctuations and to extract long-term trends from noisy time series. There are many variants of MA, including a simple moving average (SMA) and an exponential moving average (EMA). Among these variants, EMA is preferred when MA needs to quickly respond to the recent data because the past ones decay exponentially across time. In addition, unlike SMA, EMA does not require redundant computation caused by window shifting and contains the full past history of a time series due to recursive formulation. The value of EMA $\mu_{t}$ at time step $t$ is calculated by
\begin{equation}
\label{EMA}
    \mu_{t} = \alpha \cdot x_{t} + (1-\alpha) \cdot \mu_{t-1} 
\end{equation}
where $x_{t}$ is the current input value and $\alpha$ is a constant decay factor or smoothing factor between 0 and 1.

Detrending is a method for removing of a slowly changing component, called a trend, in order to make stationary time series. We think that detrending can be applied to RNNs for the elimination of internal covariate shift over time.
We notice that the definition of EMA in (\ref{EMA}) is the same as in (\ref{GRU:hidden}) for the hidden state $\mathbf{h}$ of GRU, while the decay factor $\alpha$ is continuously changing at each time step as shown in (\ref{GRU:updateGate}), rather than fixed. By considering the hidden state $\mathbf{h}$ as the trend of the candidate hidden state $\tilde{\textbf{h}}$, we can bring the mechanism of detrending to GRU for temporal normalization as follows:
\begin{equation}
    \mathbf{y}_{t} = \tilde{\textbf{h}}_{t} - \mathbf{h}_{t} 
\end{equation}
where $\mathbf{y}_{t}$ is the detrended output at time step $t$. The detrended output $\mathbf{y}_{t}$ is fed into the next layer as an input. See the schematic in Fig. \ref{fig:Model}.

We call the proposed detrending method \textit{adaptive detrending} (AD) to differentiate from the conventional detrending methods that follow a pre-defined setting to estimate a trend. 
We can get several benefits by using AD. 
First, AD requires negligible computational and memory requirements because the statistics estimation is already included as a part of the GRU computation. 
Second, AD is fully differentiable. Differentiable is necessary to normalize activations of neural networks because a statistics estimation and normalization must be included in the gradient descent optimization to prevent the model explosion, as mentioned by Ioffe and Szegedy \cite{Ioffe:2015}.
Third, AD estimates automatically any shape of a trend by adapting a decay factor at each time step and each sample. Hence, we do not need to worry about how to define a trend fitting function (such as a linear, polynomial, and moving average), or how to set the parameters of a trend fitting function (such as the window size of a moving average). Furthermore, an estimated trend by AD works as a control for the degree of detrending over time and samples. It is crucial because a fixed degree of detrending might lose some informative frequency components, which can be changed over time and samples. 
Finally, AD achieves both a sample-wise and neuron-wise normalization by using the time domain for normalization. Unlike BN, the sample-wise normalization of AD removes the dependencies between samples in a mini-batch, so that AD can be applied to RNNs without constraints. Also, unlike LN, the neuron-wise normalization of AD makes AD to be applied to a network regardless whether neurons have similar statistics (e.g., MLP) or not (e.g., CNNs).

Note that, unlike the spatial normalization methods, we do not use an affine transformation after AD because AD itself acts as a temporal normalizer following an affine transformation with the gain $\gamma$ and bias $\beta$, which are changed over time and samples, for each neuron.

\subsection{Convolutional Gated Recurrent Unit}
Convolutional gated recurrent unit (ConvGRU) is naturally extended from GRU by following the convolutional property of CNNs defined as follows:
\begin{equation}
\label{ConvGRU:resetGate}
    \mathbf{r}_{t} = \sigma(\mathbf{W}_{r} * \mathbf{x}_{t} + \mathbf{U}_{r} * \mathbf{h}_{t-1} + \mathbf{b}_{h}) 
\end{equation}
\begin{equation}
\label{ConvGRU:updateGate}
    \mathbf{z}_{t} = \sigma(\mathbf{W}_{z} * \mathbf{x}_{t} + \mathbf{U}_{z} * \mathbf{h}_{t-1} + \mathbf{b}_{z})
\end{equation}
\begin{equation}
\label{ConvGRU:newHidden}
    \tilde{\textbf{h}}_{t} = \tanh(\mathbf{W}_{h} * \mathbf{x}_{t} + \mathbf{r}_{t} \odot \mathbf{U}_{h} * \mathbf{h}_{t-1} + \mathbf{b}_{h})
\end{equation}
\begin{equation}
\label{ConvGRU:hidden}
    \mathbf{h}_{t} = \mathbf{z}_{t} \odot \tilde{\textbf{h}}_{t} + (1-\mathbf{z}_{t}) \odot \mathbf{h}_{t-1} 
\end{equation}
where $*$ is a convolution operation. The key difference between ConvGRU and GRU is that ConvGRU preserves the spatial topology because of the convolution operation on 2D feature maps with 2D weight kernels. Furthermore, ConvGRU drastically reduces the number of parameters compared with GRU when it directly applied on the spatial domain. Both the spatial normalization methods and AD can be applied to ConvGRU in the same manner as GRU.

\section{Experiments}

We are focused on contextual video recognition to show that (1) ConvRNNs, especially for ConvGRU, can successfully recognize a video by extracting spatio-temporal features at multiple scales and (2) the proposed method can offer the significant training speed up to ConvGRU.
We do not use the popular action recognition datasets including UCF-101 \cite{UCF101} and HMDB-51 \cite{HMDB51} because feed-forward networks have already performed well on these datasets with only short-term information processing \cite{Simonyan:2014,Tran:2015}. For example, it would be very easy to categorize a video of showing a person playing a guitar without extracting temporal profile in the video but by simply categorizing an object of guitar. Rather, we use two video datasets\footnote{\urlwofont{https://github.com/haanvid/CL1AD/releases}}\footnote{\urlwofont{https://github.com/haanvid/CL2AD/releases}} proposed by Lee et al. \cite{Lee:2016} for contextual recognition required both temporal as well as spatial information. In the experiments, we compare ConvGRU with (1) spatial CNN receiving a single RGB image as input and (2) convolutional 3D (C3D) \cite{Tran:2015} receiving a clip of 16 frames as input for short-term information processing to show that long-term information is crucial for contextual recognition, which cannot be shown on UCF-101 and HMDB-51. Details about the network configuration, training, and evaluation protocol for spatial CNN and C3D are provided in Appendix \ref{appx:CNN} and \ref{appx:C3D}, respectively.

\subsection{Implementation Details}

\subsubsection{Architecture}

\begin{table*}[t]
\centering
\caption{Network configuration. The filter has four dimensions (height$\times$width$\times$input channels$\times$output channels) for convolutional and fully-connected layers or two dimensions (height$\times$width) for pooling layers, and both the stride and pad have two dimensions (height$\times$width)}
\label{tab:architecture}
\begin{tabular}{|c|l|l|c|c|c|c|c|}
\hline
\multirow{2}{*}{Layer} & \multicolumn{1}{c|}{\multirow{2}{*}{Type}} & \multicolumn{3}{c|}{Forward}                        & \multicolumn{3}{c|}{Recurrent}                                       \\ \cline{3-8} 
                       & \multicolumn{1}{c|}{}                      & \multicolumn{1}{c|}{Filter}          & Stride & Pad & Filter                                                & Stride & Pad \\ \hline\hline
1                      & Conv (ReLU)                                & 7$\times$7$\times$3$\times$32        & 3$\times$3      & 0$\times$0   & -                                                     & -      & -   \\ \hline
2                      & Max                                        & 3$\times$3                           & 3$\times$3      & 0$\times$0   & -                                                     & -      & -   \\ \hline
3                      & ConvGRU                                    & 3$\times$3$\times$32$\times$64       & 1$\times$1      & 1$\times$1   & \multicolumn{1}{l|}{3$\times$3$\times$64$\times$64}   & 1$\times$1      & 1$\times$1   \\ \hline
4                      & Max                                        & 2$\times$2                           & 2$\times$2      & 0$\times$0   & -                                                     & -      & -   \\ \hline
5                      & ConvGRU                                    & 3$\times$3$\times$64$\times$128      & 1$\times$1      & 1$\times$1   & \multicolumn{1}{l|}{3$\times$3$\times$128$\times$128} & 1$\times$1      & 1$\times$1   \\ \hline
6                      & Global Avg                                 & 6$\times$6                           & -      & -   & -                                                     & -      & -   \\ \hline
\multirow{3}{*}{7}     & FC (Softmax)                               & 1$\times$1$\times$128$\times$$C_{1}$ & -      & -   & -                                                     & -      & -   \\ \cline{2-8} 
                       & \multicolumn{7}{c|}{\setstackgap{S}{.7pt}\Shortstack{. . .}}                                                                                                  \\ \cline{2-8} 
                       & FC (Softmax)                               & 1$\times$1$\times$128$\times$$C_{N}$ & -      & -   & -                                                     & -      & -   \\ \hline
\end{tabular}
\end{table*}
The networks consist of one convolutional (Conv), two convolutional gated recurrent unit (ConvGRU), two max pooling (Max), one global average pooling (Global Avg), and one fully-connected (FC) layers. The bottom layer is a convolutional layer for spatial dimension reduction having 32 feature maps with 7$\times$7 kernels and 3$\times$3 stride, and followed by a rectified linear unit (ReLU) activation function. Then, two ConvGRU layers are stacked having 64 and 128 feature maps, respectively, with 3$\times$3 kernels and 1$\times$1 padding for both the forward $\mathbf{W}$ and recurrent $\mathbf{U}$ paths. The two max pooling layers are located in between the three layers (including one convolutional and two ConvGRU layers) with subsampling factors of 3$\times$3 and 2$\times$2 for the first and second max pooling, respectively. A global average pooling layer \cite{Lin:2013} follows the last ConvGRU layer to vectorize the all feature maps, and the vectorized feature maps are fed into a fully-connected layer with a softmax activation function for each category. When there are $N$ categories in a dataset, the $n$-th category has $C_{n}$ classes. The description of the network configuration is summarized in Table \ref{tab:architecture}.


\subsubsection{Training}
\label{sec:training}
The networks are trained by the mini-batch stochastic gradient descent (SGD) with Nesterov's accelerated gradient (NAG) \cite{Nesterov:1983,Nesterov:2004}, setting the momentum coefficient $\mu$ to 0.9, and implemented in Torch7 \cite{torch}. The size of mini-batch is set to 8. The gradient is calculated using a back-propagation through time (BPTT) algorithm. We use the negative log likelihood loss function $\ell$ as follows:
\begin{equation}\label{eq:NLL} 
    \ell = -\sum_{c=1}^{C}p_{c}\log(\hat{p}_{c})
\end{equation}
where $C$ is the number of classes, and $p_{c}$ and $\hat{p}_{c}$ are the true and predicted probability of class $c$, respectively.
The error is only generated at the end of a training sequence to utilize the accumulated information through space and time for recognition as used by Jung et al. \cite{Jung:2015}.
The L2-norm weight decay of 0.0005 is applied while updating the network parameters in order to prevent over-fitting as used by Krizhevsky et al. \cite{Krizhevsky:2012}. Because of the exploding gradient problem, we use a gradient clipping method \cite{Pascanu:2013}, which rescales the L2-norm of the gradient to a threshold whenever the L2-norm exceeds a threshold. Here, a threshold is set to 10. The initial hidden state $\mathbf{h}_{0}$ in ConvGRU is set to 0. 




All weights are initialized using randomly selected values from a zero-mean Gaussian distribution with a standard deviation $\sigma$ setting differently depending on the experiment. 
Similar to the bias initialization trick used to solve the gradient vanishing problem of LSTM \cite{Gers:2000,Jozefowicz:2015}, the update gate biases are initialized to -2, and the remaining biases are initialized to 0 by default unless otherwise noted.
For both recurrent batch normalization (recurrent BN) \cite{Cooijmans:2016} and layer normalization (LN) \cite{Ba:2016}, the gain $\gamma$ and bias $\beta$ of each affine transformation are initialized to 1 and 0, respectively. 
However, exceptionally, when recurrent BN and LN are applied to the update gate, the bias $\beta$ of each affine transformation is initialized to -2 to follow the above bias initialization trick. Note that, we do not initialize the gain $\gamma$ to 0.1 as used by Cooijmans et al. \cite{Cooijmans:2016} because it shows worse results than when the gain $\gamma$ is initialized to 1 in the following experiments.

\subsubsection{Data Pre-processing and Augmentation} 
\label{sec:preprocessing}
The videos given to the network are rescaled to 128$\times$170 pixels and normalized the range from integer values 0 to 255 to real values -1 to 1. To reduce the over-fitting problem, we process all images in the same video by randomly sampling a 112$\times$112 region, and then randomly flipping the images horizontally with 50\% probability.

\subsubsection{Testing} 
We obtain 10 sequences for each test sequence by cropping 1 center and 4 corners, and then horizontally flipping them. The output at the end of a testing sequence is used for classification score in the same manner as the training phase. The final classification accuracy is obtained by averaging of 10 classification scores.

\subsection{Evaluation Protocol}
We divide both datasets used in the following two experiments into three splits for cross-validation. Each split contains eight subjects for training and two subjects for testing. After getting recognition accuracy of three splits, we report the average recognition accuracy over three splits.

\subsection{Object-Related Action Recognition}
A dataset for the object-related action (OA) recognition contains 900 videos in 15 object-action combination classes, which are a partial combination of four objects (`Book', `Laptop', `Bottle', and `Cup') and nine actions (`Change page', `Sweep', `Open', `Close', `Type', `Shake', `Drink', `Stir', and `Blow') rather than a full combination of them. To avoid action inference by object identification, each video in the dataset shoot with two objects; one is a target object on which an action needs to be performed by a subject, and the other is a distractor, which is not related to the current action. Each object-action combination class is performed by 10 subjects for two times with three distractors. The viewpoint and background are static. We pre-process all videos to have a fixed number of frames (50 frames) because batch normalization is difficult to handle variable length sequences in a mini-batch. All networks are initialized with a standard deviation of 0.07 and trained with a learning rate of 0.01 over 100 epochs. 

\begin{table}[t]
\caption{Accuracy comparison on OA recognition dataset}
\label{tab:exp1_accuracy}
\centering
\begin{tabular}{|l|cc|c|}
\hline
\multicolumn{1}{|c}{\multirow{2}{*}{Model}} & \multicolumn{3}{|c|}{Accuracy}                                                               \\ \cline{2-4} 
                       & Object               & Action           &        Joint               \\ \hline
Spatial CNN & 87.4\% & 64.1\% & 59.8\% \\ 
C3D   & 99.0\% &	98.0\% &	97.2\%   \\ \hline\hline
Baseline (init bias: 0) & 98.1\% &	98.0\% & 96.9\%   \\
Baseline (init bias: -2) & 98.7\% &	97.4\% & 97.0\%   \\\hline
AD (init bias: 0) & 99.3\% &	98.7\% & 98.3\%   \\
AD (init bias: -2) & 99.1\% &	98.9\% & 98.5\%   \\\hline
BN (all)  & 98.9\% &	98.9\% &	98.1\%    \\
BN (hidden) & 99.1\% &	98.3\% &	97.8\%    \\
BN (gates) & 99.1\% &	97.6\% &	97.0\%    \\\hline
LN (all)  & 98.9\% & 98.3\% & 97.6\%    \\
LN (hidden) & 98.9\% &	98.1\% & 97.4\%    \\
LN (gates) & 98.3\% &	97.6\% & 96.5\%    \\\hline
BN+AD & 99.4\% &	98.9\% &	98.5\%    \\\hline
LN+AD & 98.9\% &	99.4\% &	98.5\%  \\ \hline
\end{tabular}
\end{table}

\subsubsection{Feed-Forward Networks}
Table \ref{tab:exp1_accuracy} shows the test accuracy of the networks. The accuracy gap between spatial CNN and C3D indicates that at least short-term information needs to be processed for OA recognition. In the case of spatial CNN without using any temporal information, a distractor and the similarity between actions make difficult to recognize object and action, respectively. 

\subsubsection{Initialization of Update Gate Bias}
\begin{figure}[t]
\centering
\subfloat[]{\includegraphics[width=3in]{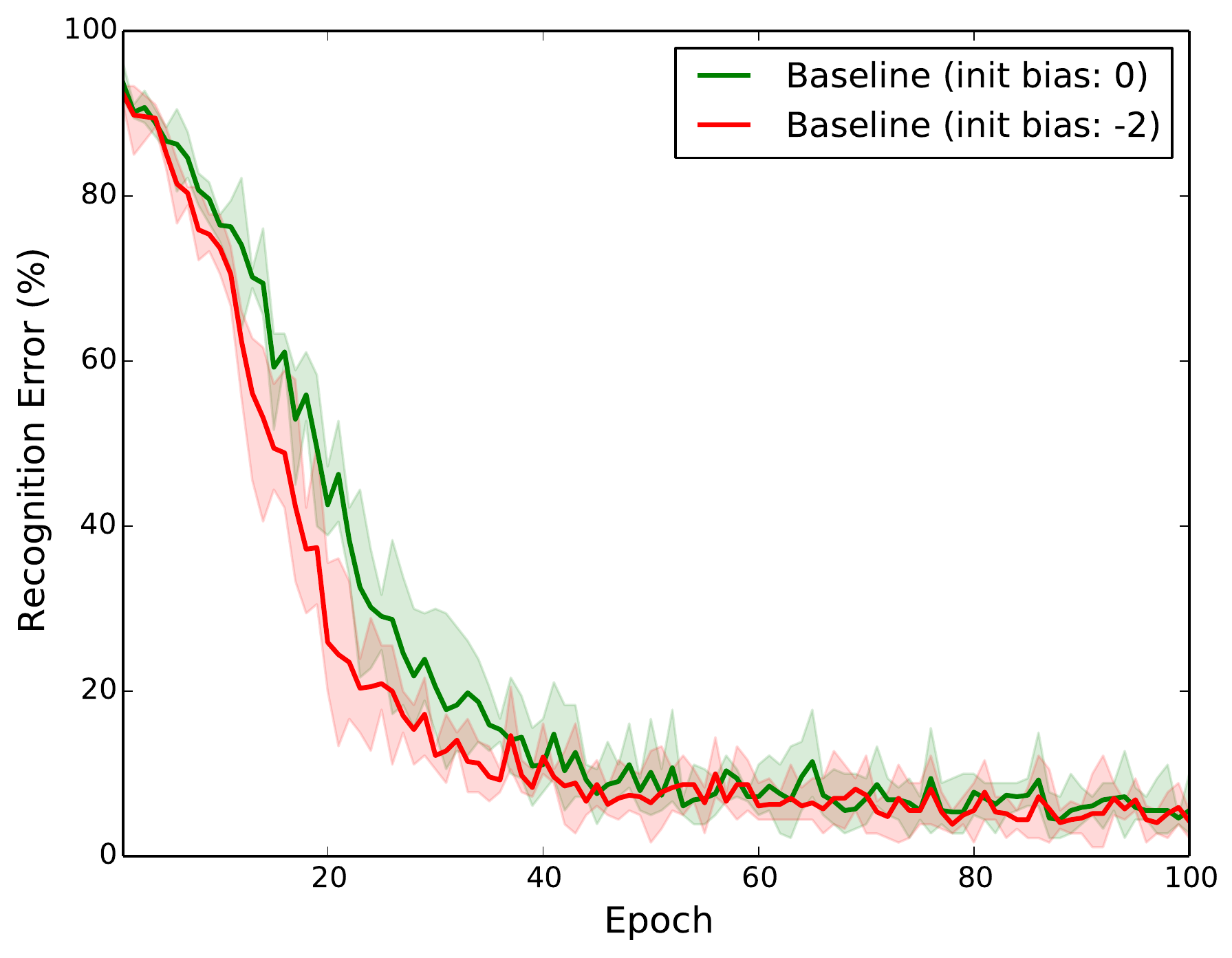}
\label{fig:initBias_Baseline}}
\hfil
\subfloat[]{\includegraphics[width=3in]{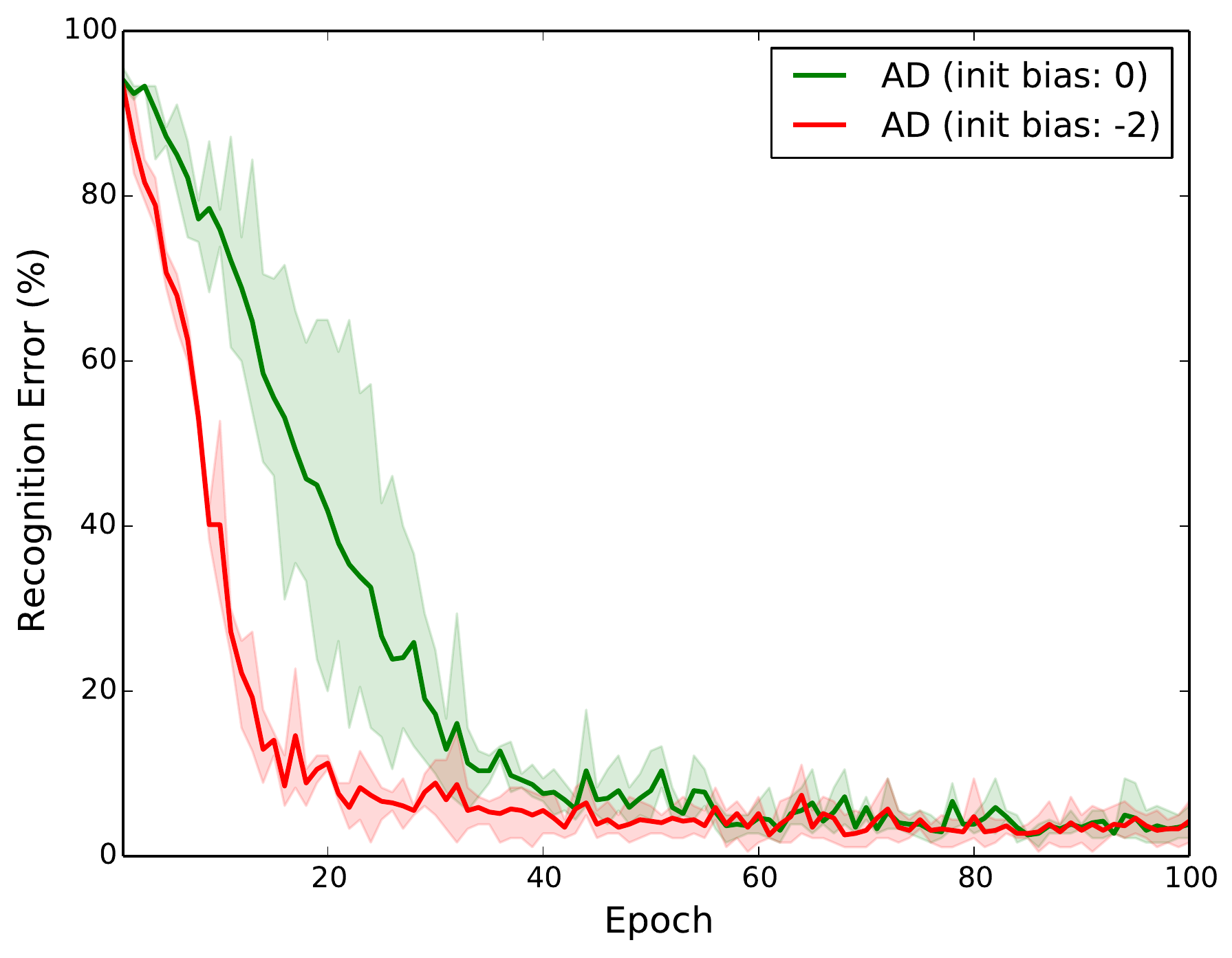}
\label{fig:initBias_AD}}
\caption{Importance of initializing the update gate bias on object-related action (OA) recognition dataset. The graphs show the comparison between two initialization strategies for the update gate bias of (a) the baseline and (b) AD. Solid lines represent the test recognition errors averaged over three splits and shaded regions represent the differences between the maximum and minimum values of three splits.}
\label{fig:initBias}
\end{figure}

\begin{figure}[t]
\centering
\includegraphics[width=3in]{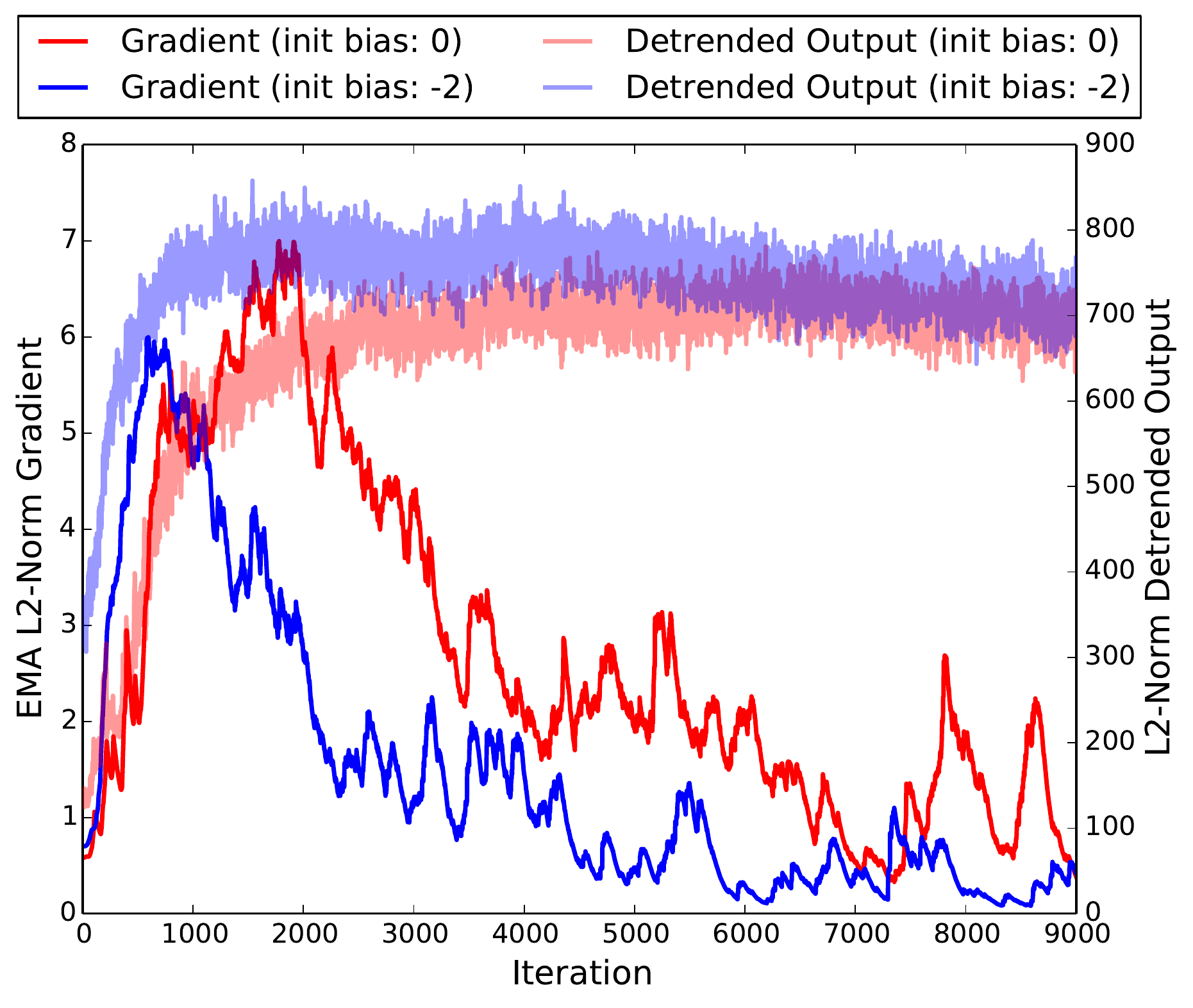}
\caption{L2-norm of the gradient (solid lines) and detrended output (semi-transparent lines) of AD versus iteration on OA recognition dataset (split 1). Each L2-norm of the gradient is smoothed by EMA with a decay factor of 0.99 for better visualization.}
\label{fig:L2-norm}
\end{figure}

Initializing the forget gate bias to a large positive value (usually 1 or 2) is a widely used trick for LSTM to prevent the gradient vanishing problem when the weights and biases of LSTM are initialized to small random numbers \cite{Gers:2000,Jozefowicz:2015}. In more detail, a random initialization sets the forget gate to be centered on 0.5, so that the bias initial information of LSTM is exponentially decaying out through time. By using the initialization trick, the performance and convergence speed of LSTM are improved, especially when long-term dependencies are crucial. In the case of GRU, initializing the update gate bias to a large negative value provides the same effect as the bias initialization trick of LSTM. 

In the experiment, we compare the convergence speeds between zero and negative, here set to -2, initial biases of the update gate on the baseline model (ConvGRU without normalization) and adaptive detrending (AD) to examine the effect of the bias initialization trick. In Fig. \ref{fig:initBias}, we observe that both the baseline and AD show the convergence speed improvement by initializing the update gate bias to -2 rather than 0, but the improvement by AD is much more significant than that of the baseline. Furthermore, when the update gate bias of AD is initialized to 0, the variance of convergence graphs trained on three different splits is larger than the others (the baseline with the zero and negative initial biases, respectively, and AD with the negative initial bias). These results indicate that (1) a random initialization causes severely slow and unstable learning problems to AD, and (2) the initialization of the update gate bias is more crucial to AD than the baseline. 

We hypothesize that the detrended output of AD initially are too small to make an enough gradient for training when the update gate bias is initialized to 0, because the hidden state (or trend) of AD closely follows an input sequence. To verify our hypothesis, we further analyze the L2-norm of the detrended output and that of the gradient for both the zero and negative bias initialization of AD during training as shown in Fig. \ref{fig:L2-norm}. Because the L2-norm of the gradient shows a high fluctuation, we smooth it using an exponential moving average (EMA) with a decay factor of 0.99. As we expected, the L2-norm of the detrended output of the zero bias initialization is considerably smaller than that of the negative bias initialization during the initial phase of learning. As a result, BPTT can not generate an enough gradient for training, so that the zero bias initialization slows down the training of the network compared to the negative bias initialization.
Hence, from now on, the bias of the update gate will be initialized to -2.

\subsubsection{Overhead Reduction for Spatial Normalization}
\begin{figure}[t]
\centering
\subfloat[]{\includegraphics[width=3in]{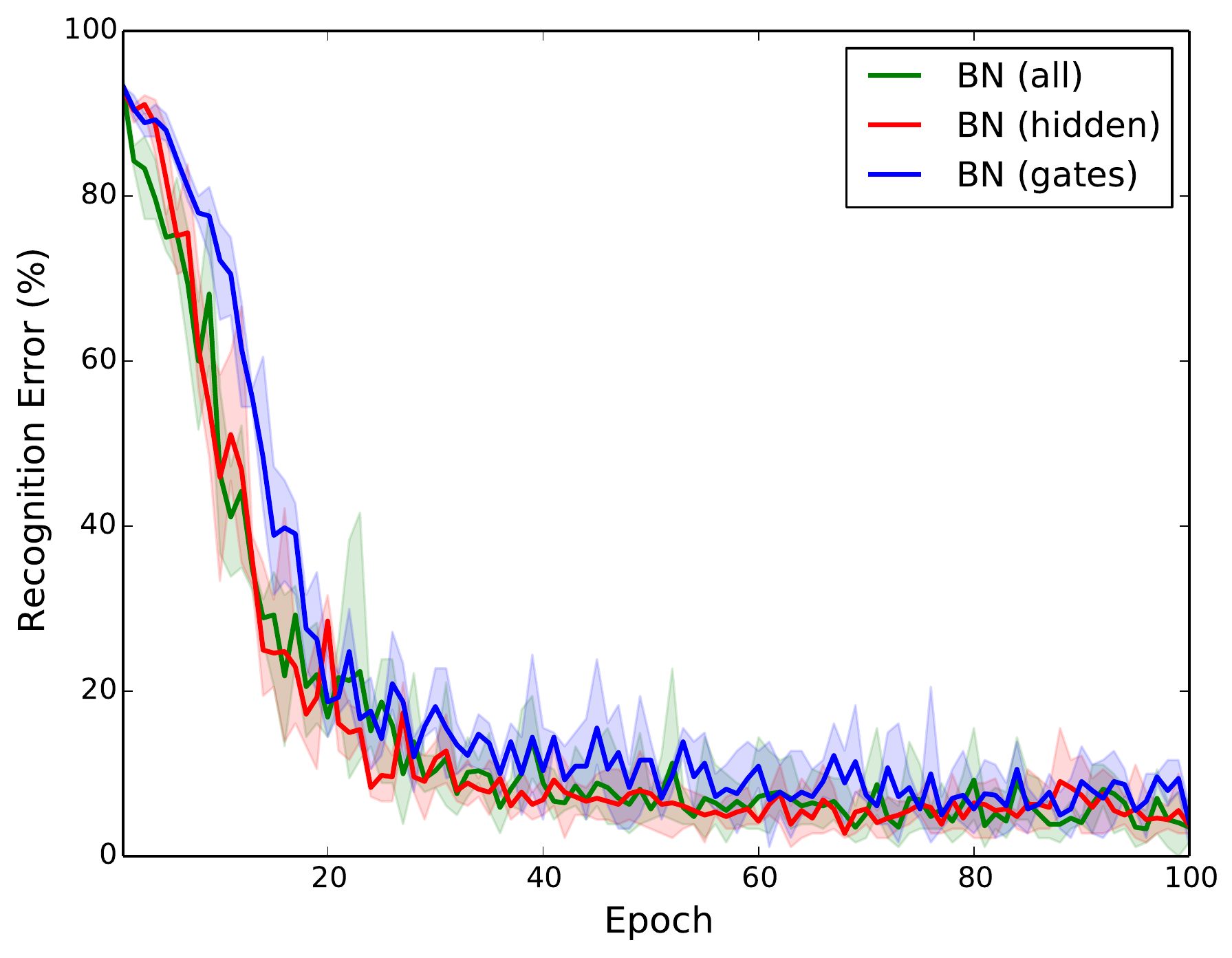}
\label{fig:BN}}
\hfil
\subfloat[]{\includegraphics[width=3in]{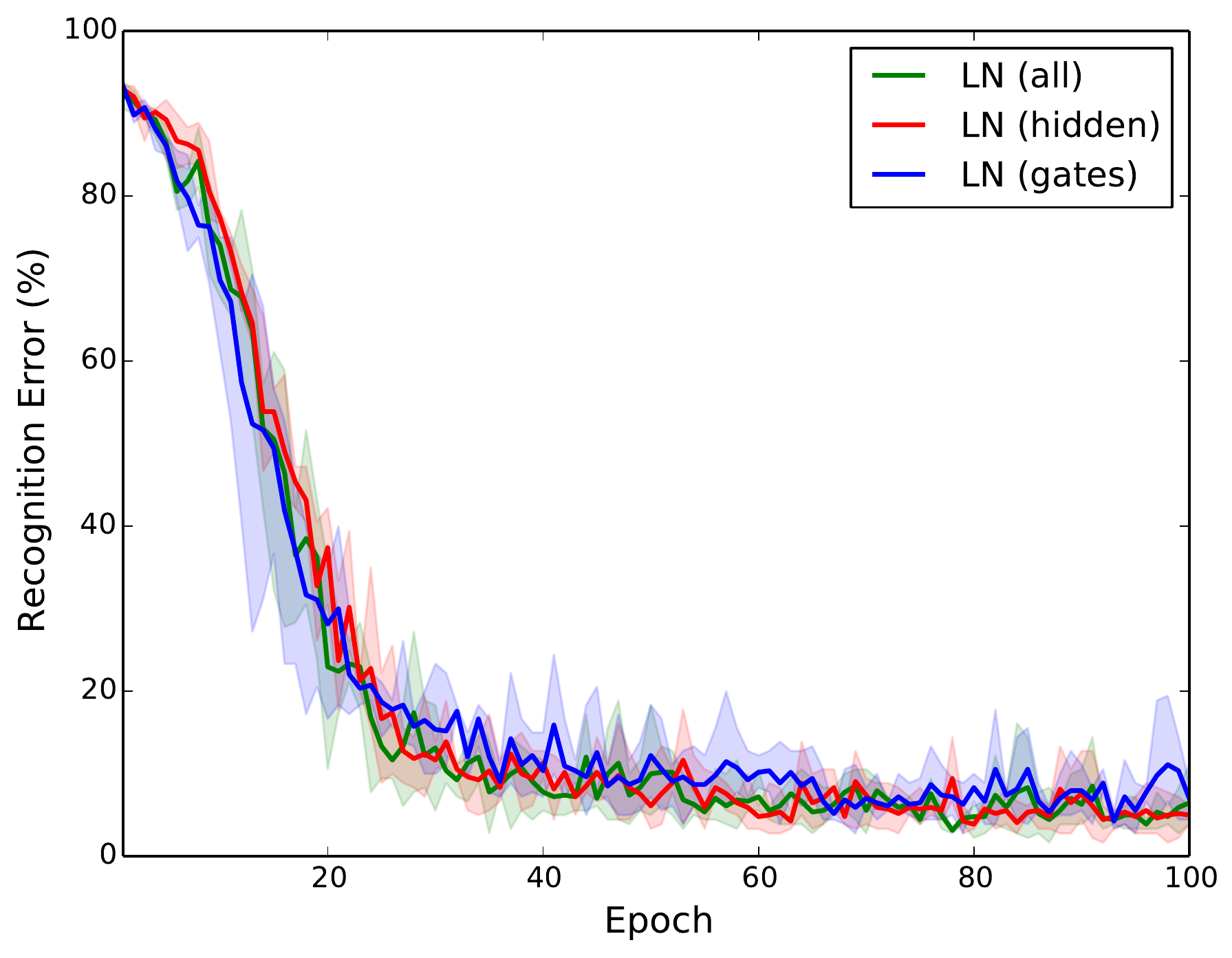}
\label{fig:LN}}
\caption{Effect of spatial normalization according to their location on OA recognition dataset. (a) Batch normalization (BN). (b) Layer normalization (LN). For `hidden', the bias of the update gate is initialized to -2. Otherwise, the bias of an affine transformation for the update gate is initialized to -2. Solid lines and shaded regions are the same as that in Fig. \ref{fig:initBias}.}
\label{fig:SN}
\end{figure}

Although the spatial normalization methods can provide the acceleration of deep neural network training, the computation cost and memory consumption are additionally required to estimate statistics and to normalize, which is called an overhead. Given the benefits of the spatial normalization methods, the overhead of normalization might be endurable in most of the neural networks. However, when the spatial normalization methods are applied to ConvRNNs, the prohibitively slow training and memory overflow problems often occur because the computational cost and memory consumption of naive ConvRNNs are already too large.


As we explained in Section \ref{sec:norm}, recurrent BN (hereafter, we simply abbreviate it to BN by taking `recurrent' out) and LN normalize the three input distributions of the following: the candidate hidden state, reset gate, and update gate of GRU.
However, we consider that all three normalizations might not contribute equally to the improvement achieved by BN and LN. If this assumption is true, we can eliminate less important normalizations to alleviate the overhead while minimizing the performance and training speed degradation.
Hence, we investigate the effects of normalization depending on where BN and LN are employed, as follows: the candidate hidden state (`hidden'); reset and update gates (`gates'); and candidate hidden state, and reset and update gates (`all').

Fig. \ref{fig:SN} shows the convergence speeds of `hidden', `gates', and `all' of BN and LN. Compared with `all', we observe that `gates' slows down marginally the convergence speed of `BN' and converges to worse local optima for both BN and LN, which matches the performance degradation of `gates' (around 1.0\%) reported in Table \ref{tab:exp1_accuracy}.
However, in the case of `hidden' for BN and LN, although the performance is slightly decreased (around 0.2\%), the convergence speed is similar to that of `all'. The amount of performance degradation of `hidden' is acceptable because it requires only one third of the original overhead for `all'. These results indicate that the normalization of the candidate hidden state plays the most important role of the three normalizations. Hence, we will apply BN and LN only to the candidate hidden state for the rest of the experiments.

\subsubsection{Adaptive Detrending versus Spatial Normalization}
\label{AD_vs_SN}

\begin{figure}[t]
\centering
\subfloat[]{\includegraphics[width=3in]{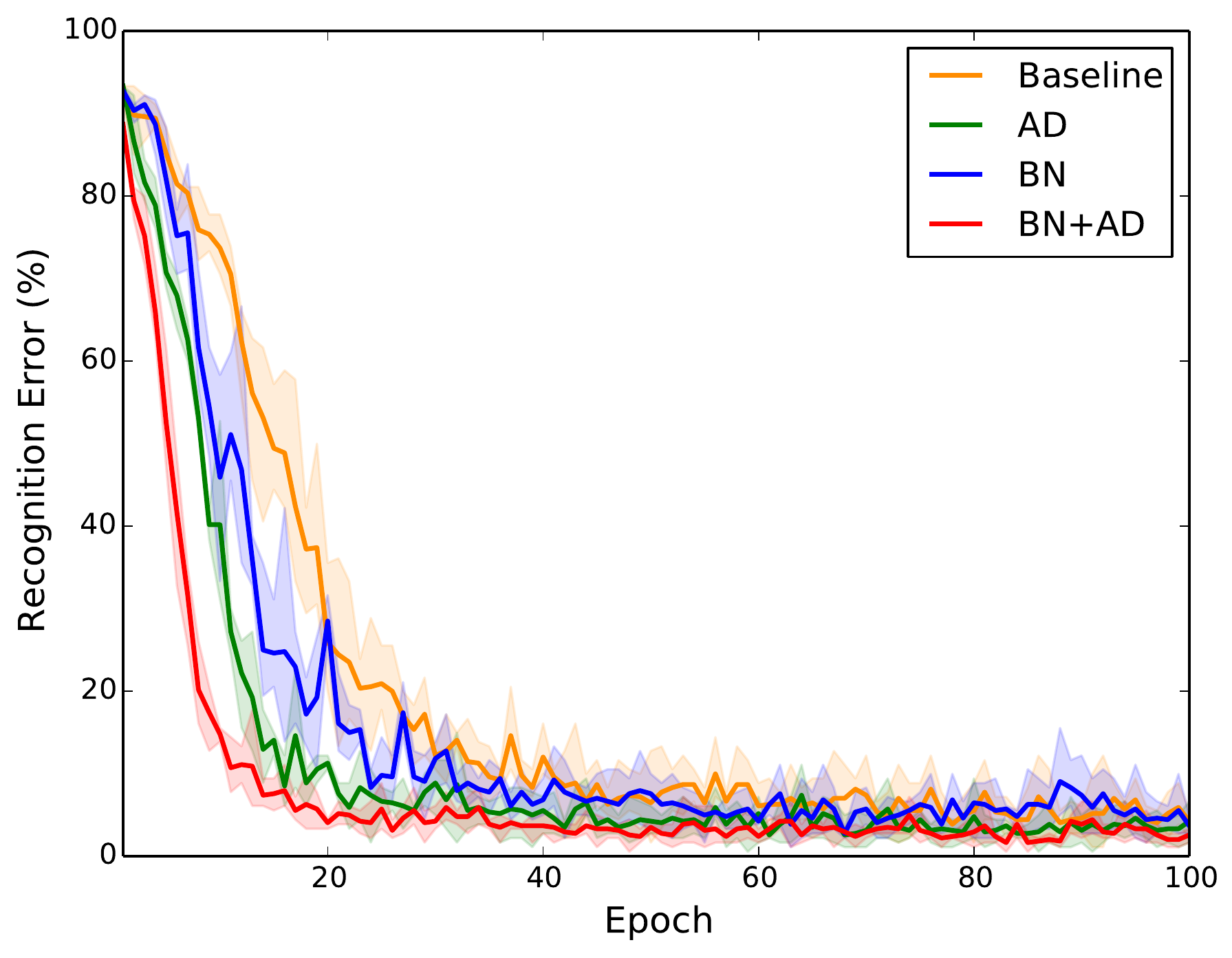}
\label{fig_second_case}}
\hfil
\subfloat[]{\includegraphics[width=3in]{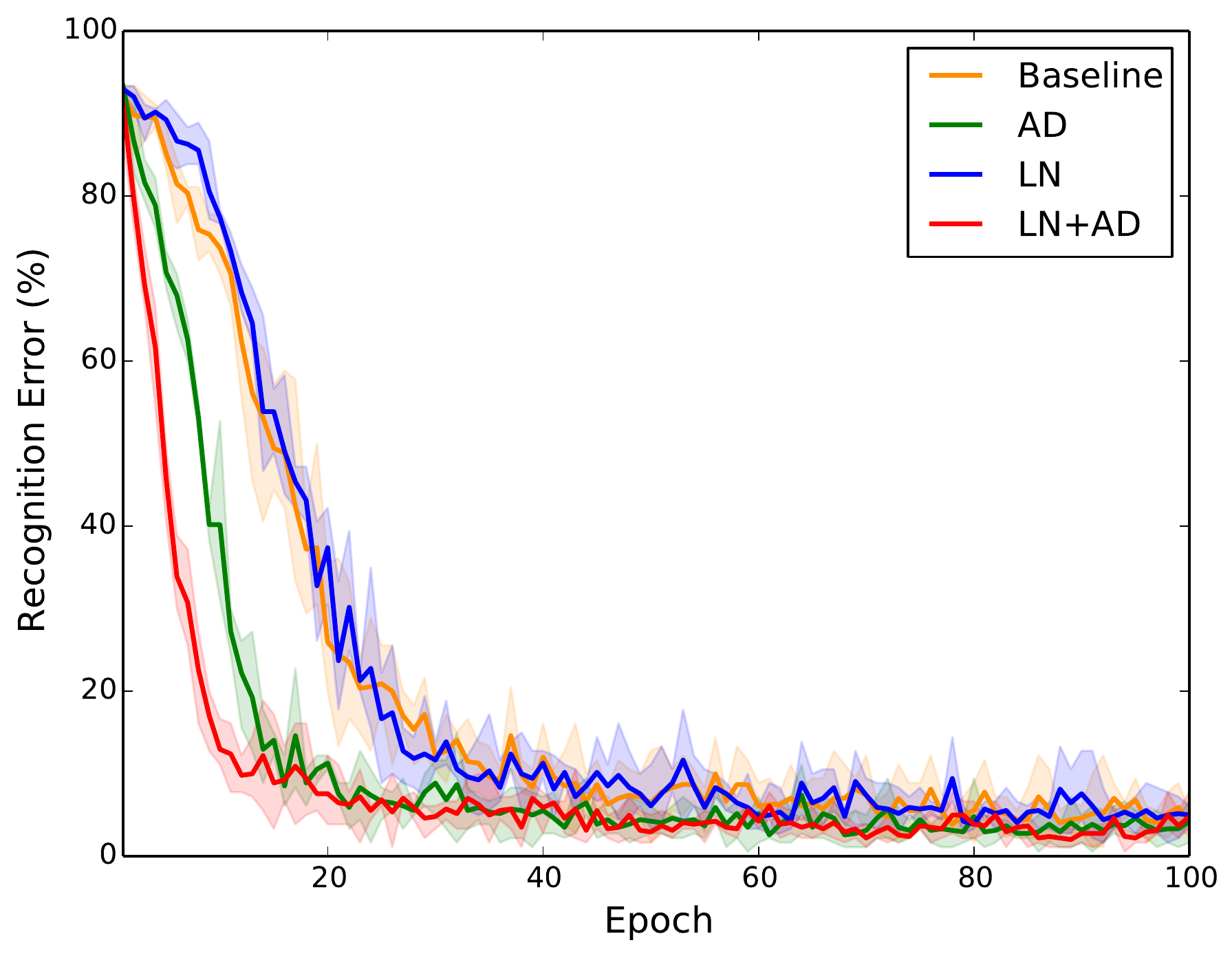}
\label{fig_first_case}}
\caption{Graph of the test recognition error averaged over three split versus training epochs on OA recognition dataset. (a) Comparison of the baseline, AD, BN, and BN+AD. (b) Comparison of the baseline, AD, LN, and LN+AD. The spatial normalization methods (BN and LN) are only applied to the candidate hidden state. The bias of the update gate is initialized to -2. Solid lines and shaded regions are the same as that in Fig. \ref{fig:initBias}.}
\label{fig:comparison}
\end{figure}

\begin{figure*}[t]
\centering
\subfloat[]{\includegraphics[width=3.5in]{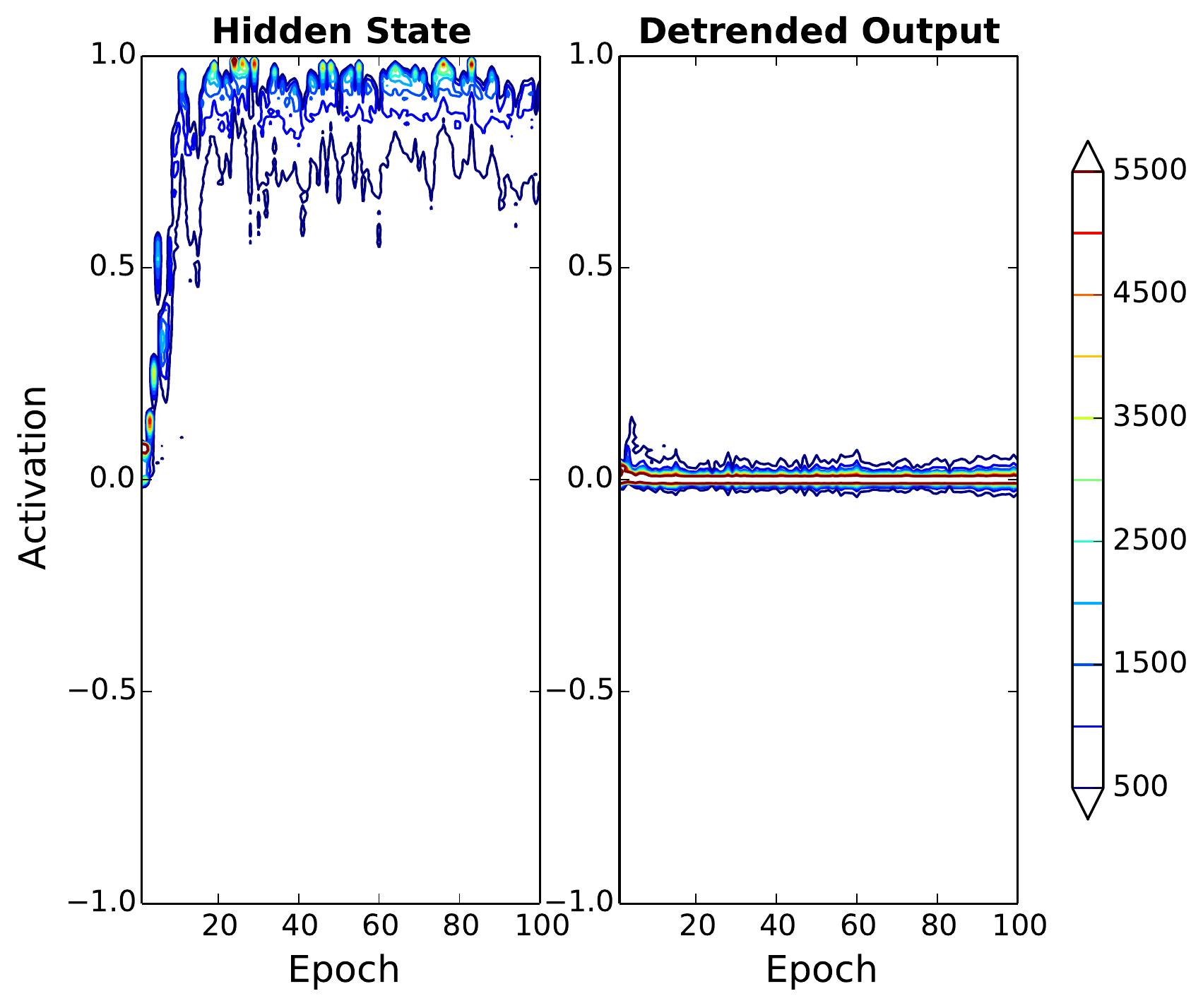}
\label{fig:}}
\hfil
\subfloat[]{\includegraphics[width=3.5in]{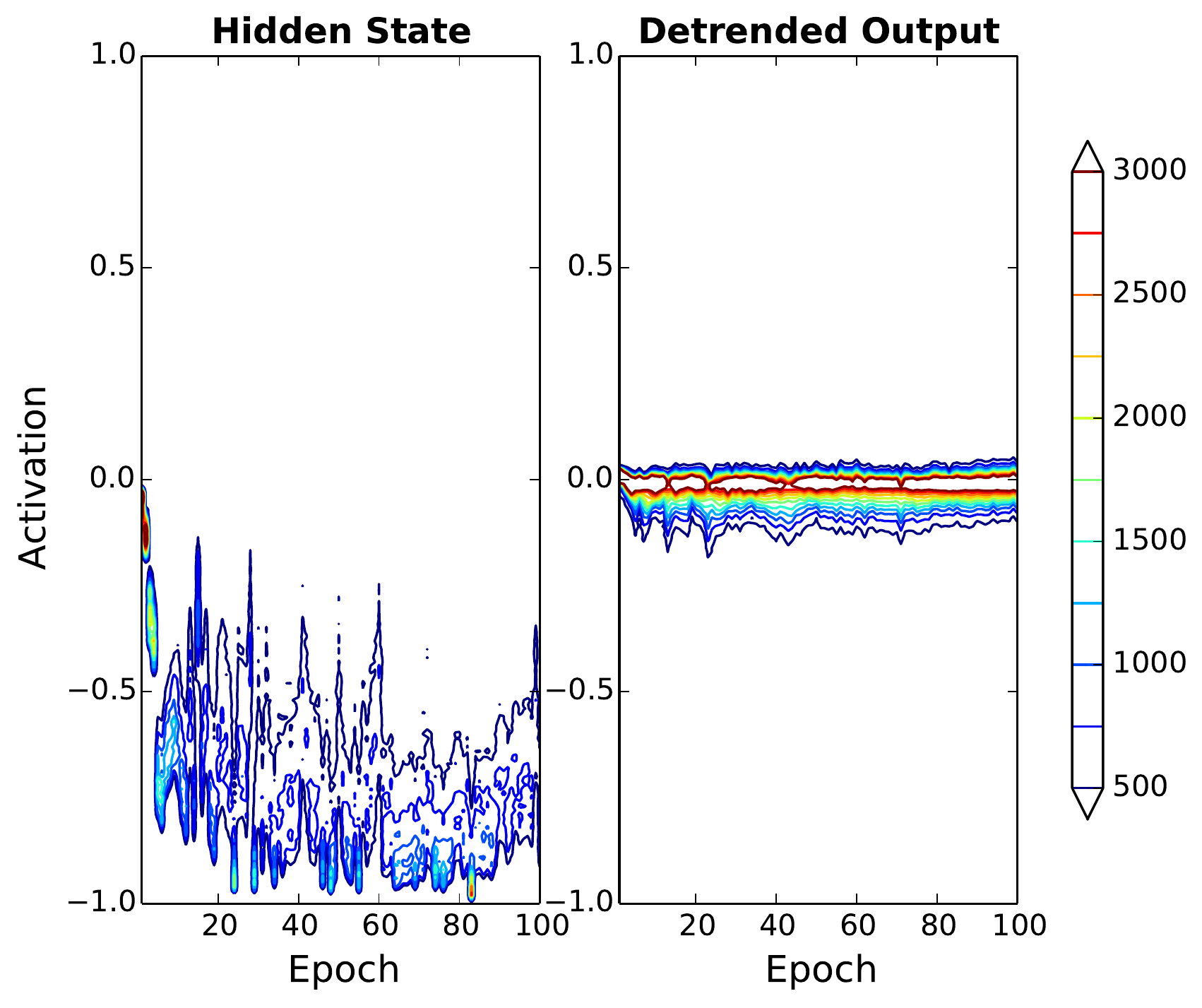}
\label{fig:}}
\caption{Visualization of internal covariate shift reduction by adaptive detrending for a neuron. Each contour map shows the distribution change from 1 to 100 epochs. Note that, the maximum contour level and interval of (a) and (b) are set differently for better visualization. Each distribution is approximated by a histogram of a single neuron's activations in the first ConvGRU layer for 720 videos of eight training subjects on OA recognition dataset (split 1). A histogram has bins with an interval of 0.01 from -1 to 1. Two selected neurons are shown in (a) and (b). Left panel: The distribution change of the hidden state. Right panel: The distribution change of the detrended output.}
\label{fig:ICS_reduction}
\end{figure*}

\begin{figure}[t]
\centering
\subfloat[]{\includegraphics[width=2.5in]{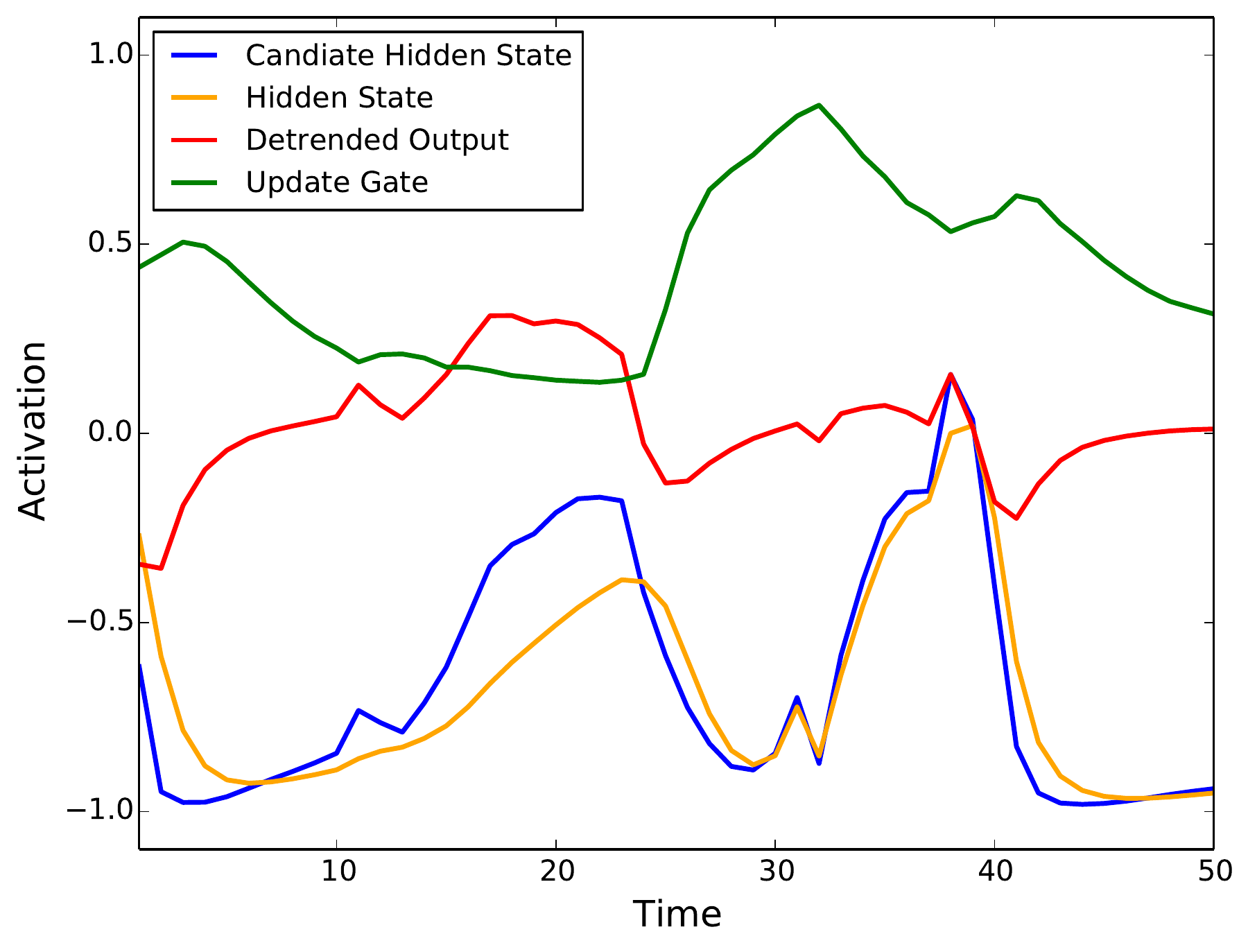}
\label{fig_second_case}}
\hfil
\subfloat[]{\includegraphics[width=2.5in]{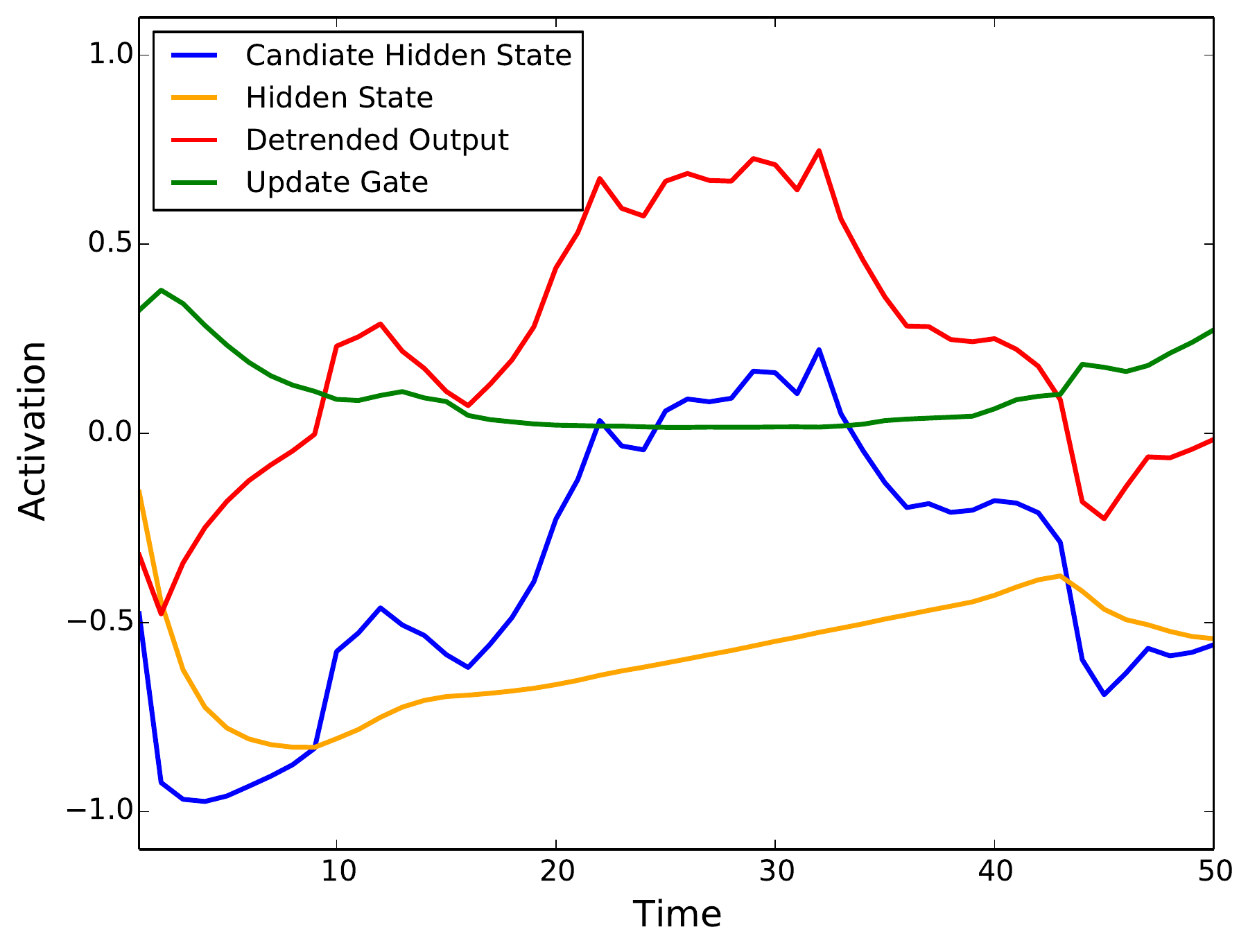}
\label{figgg}}
\hfil
\subfloat[]{\includegraphics[width=2.5in]{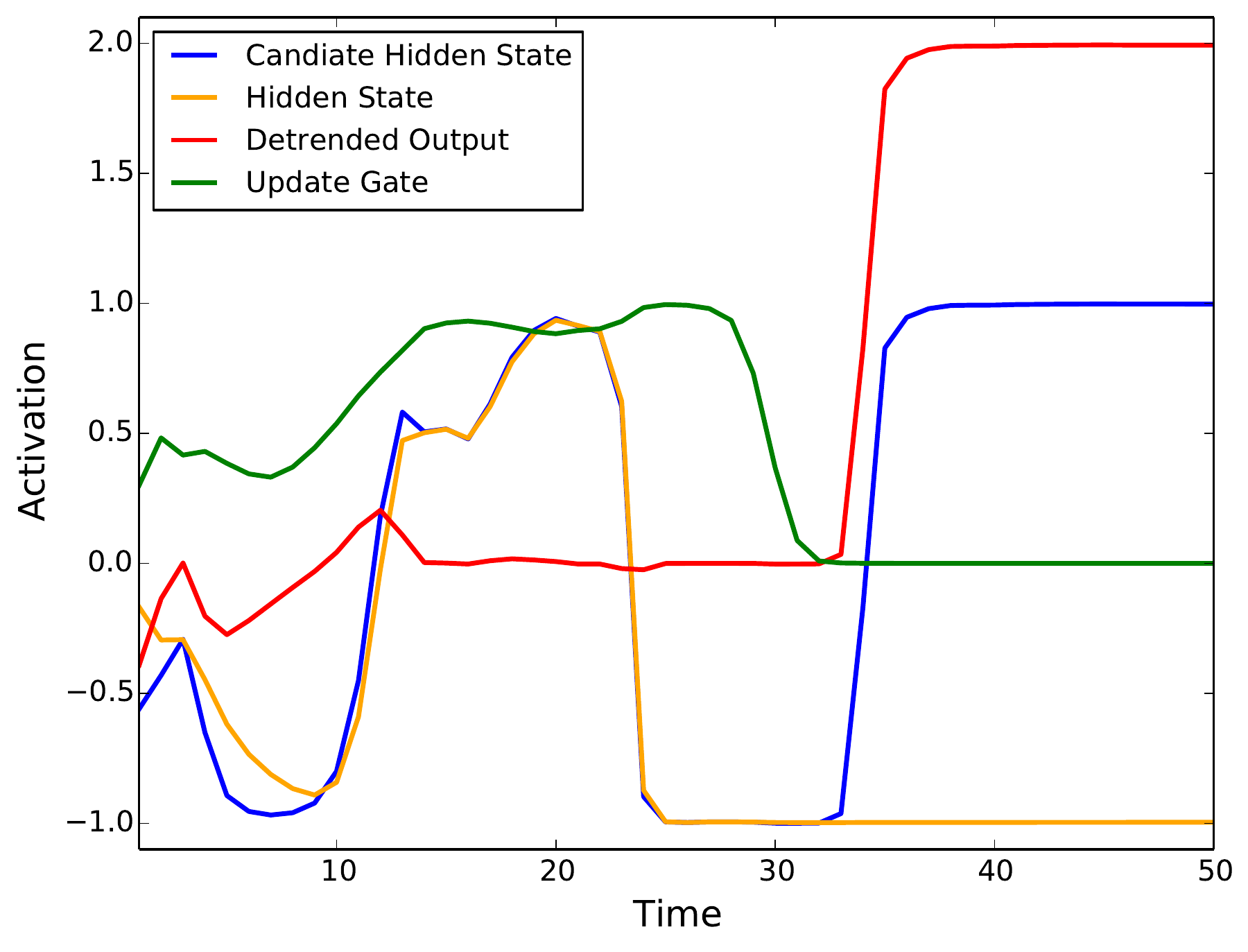}
\label{fig_first_case}}
\caption{Visualization of adaptive detrending for a neuron. Time series are obtained from a single neuron in the second ConvGRU layer while receiving a training video, after training is finished on OA recognition dataset (split 1). Two selected neurons are shown in (a)-(b) and (c). Note that, (a)-(b) are the same neuron, but receive different videos.}
\label{fig:AD}
\end{figure}

Since the spatial normalization methods have shown a speed-up and performance improvement in many different tasks \cite{Cooijmans:2016,Ba:2016}, we expect that the same benefits will be acquired by BN and LN in OA recognition task. As we expected, both BN and LN increase the recognition accuracy than the baseline, and BN accelerates the convergence speed as shown in Fig. \ref{fig:comparison} and Table \ref{tab:exp1_accuracy}. However, LN does not offer any speed-up in ConvGRU. 
Because of the statistics estimation error caused by an incorrect initial assumption, LN has already been known that it offers a speed-up over the baseline but underperforms BN when LN applied to CNNs as we explained in Section \ref{sec:LN}. Surprisingly, LN for ConvGRU is even worse than LN for CNNs. We hypothesize that the statistics estimation error of LN for CNNs is accumulated through time in the case of ConvRNNs. Hence, LN shows much worse results when applied to ConvGRU than CNNs. 

In Fig. \ref{fig:comparison} and Table \ref{tab:exp1_accuracy}, the comparison between the baseline and AD shows that AD significantly improves both the convergence speed and recognition accuracy. Furthermore, the improvement by using AD is even more than that of BN and LN. These results imply that the time domain is more critical than the spatial domain when the normalization scheme is applied to RNNs.

Now, we visualize internal activations of the network in order to investigate the mechanisms of AD in a qualitative manner. First, we analyze the reduction of internal covariate shift by AD. Fig. \ref{fig:ICS_reduction} shows the distribution change of the hidden state and that of the detrended output for a single neuron in the first ConvGRU layer over epochs. The distributions of the hidden state drastically change over epochs in terms of the mean and variance, which represents that internal covariate shift is occurred. However, the distributions of the detrended output are much more stable over epochs, which represents that AD successfully reduces internal covariate shift. These results indicate that eliminating internal covariate shift is closely related to the acceleration of convergence speed as mentioned by Ioffe and Szegedy \cite{Ioffe:2015}.

Next, we analyze how AD actually works over time in a single neuron level. In Fig. \ref{fig:AD}, we plot four time series of the following: the candidate hidden state, hidden state, update gate, and detrended output neurons in the second ConvGRU layer. From the perspective of detrending, the candidate hidden state, hidden state, and update gate can be considered to an input, trend, and decay factor, respectively. In Fig. \ref{fig:AD}(A), we observe that the trend is successfully estimated and removed from the input to generate the detrended output in the neuron. Unlike the conventional detrending methods, AD automatically controls the degree of detrending by changing the decay factor over time. Specifically, AD removes both a low- and high-frequency components or only a low-frequency components as the decay factor increases or decreases. Furthermore, Fig. \ref{fig:AD}(A) and (B) shows that AD works very differently, even in the same neuron, depending on an input. These results suggest that AD can decide which information should be maintained or removed over time and samples. Note that, a learnable affine transformation of BN and LN allows to control the degree of normalization over neurons, feature maps, or layers, but neither time nor samples. 

Interestingly, Fig. \ref{fig:AD}(C) provides the evidence why AD improves the generalization of the network. The detrended output is almost flat at zero until around 30 time steps because of the high decay factor, but then suddenly the decay factor converges to zero. Hence, for the remaining time steps, the trend is fixed to -1 while the input increases rapidly from -1 to 1; the detrend output becomes 2 (1-(-1)=2) because it is the subtraction of the trend from input. If we assume that this sudden increase in the input after 30 time steps is crucial for correct classification, AD increases the discriminability between classes by enhancing true class-related information while removing others. Note that, the detrended output in the second ConvGRU layer is directly given to the fully-connected layer. Therefore, we argue that a control for the degree of detrending by AD works not only just for reducing internal covariate shift over time, but also for better classification performance.

\subsubsection{Synergy between Adaptive Detrending and Spatial Normalization}

Because (1) each normalization in the spatial or time domains is proven to be beneficial for the training of ConvGRU in Section \ref{AD_vs_SN} and (2) two domains are not overlapped each other, each of these improvements might be combined by applying AD together with the spatial normalization methods.
Fig. \ref{fig:comparison} clearly shows that AD collaborated with BN and LN (abbreviated as BN+AD and LN+AD, respectively) further improve the convergence speed than BN, LN, or AD used alone. 
These results empirically verify our hypothesis that utilizing the time domain as well as the spatial domain for normalization will generate a strong synergy. 

Furthermore, AD can solve the difficulty of applying LN to CNNs and their variants, as described in Section \ref{AD_vs_SN}. More specifically, the neuron-wise normalization of AD, which is naturally acquired by using the time domain, overcomes the limitation of LN.
After the activations of ConvGRU having different statistics over feature maps are normalized (or detrended) by AD in a neuron-wise manner, the detrended activations having similar statistics satisfy the assumption of LN. That is why the improvement from LN to LN+AD achieved by the temporal and neuron-wise normalization is more than the improvement from BN to BN+AD achieved by the temporal normalization alone.

\subsection{Object-Related Action with Modifier Recognition}
By extending OA recognition experiment, we further examine the proposed method on the object-related action with modifier (OA-M) recognition experiment. A dataset for OA-M recognition consists of 840 videos in 42 object-action-modifier combination classes by partially combining four objects (`Box', `Book', `Cup', and `Spray'), four actions (`Move', `Touch', `Drag', and `Sweep'), and six modifiers (`To Left', `To Right', `To Front', `One Time', `Two Times', and `Three Times'). Each object-action-modifier combination class is performed by 10 subjects for two times with a randomly selected distractor. The viewpoint and background are static. Compared with OA recognition, OA-M recognition is a more complex task because adding the modifier category provides a large number of combination classes, and modifier recognition requires long-term or contextual information. For example, a network should wait until a video is finished to discriminate between modifiers `One Times' and `Two Times'. Unlike OA recognition experiment, we directly use raw videos sampled at a frame rate of 15 \textit{fps} without frame length normalization because it might lose temporal information. The maximum length of the sampled sequences is 117, which is more than two times longer than 50 frames used in OA recognition. All networks are initialized with a standard deviation of 0.05 and trained with a learning rate of 0.005 over 200 epochs. Because training sequences in a mini-batch have different lengths, each output gradient of training sequences is linearly weighted by the division of the maximum sequence length by each sequence length.

\begin{figure}[t]
\begin{center}
\epsfig{file=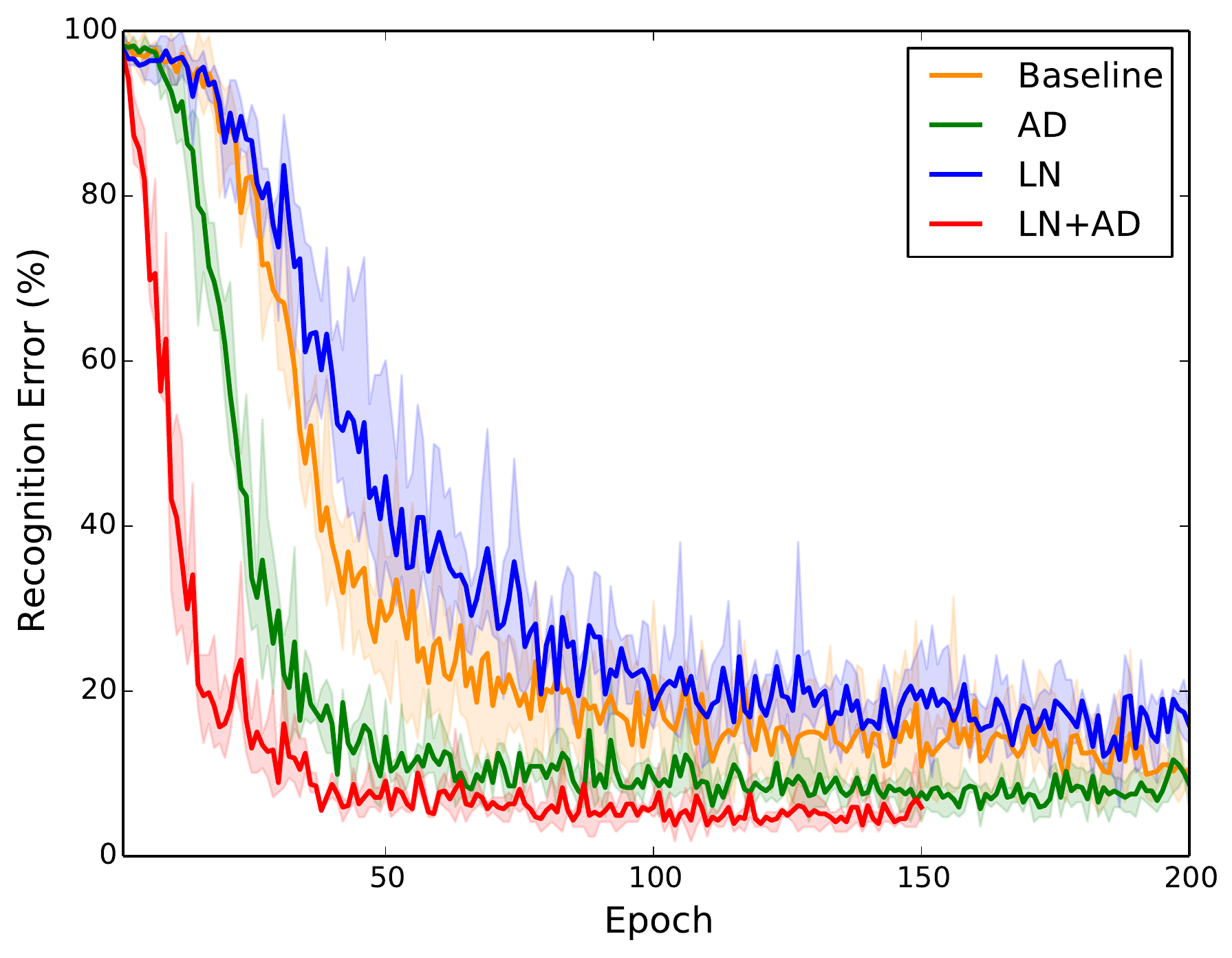, width=3in}
\caption{Graph of the test recognition error averaged over three split versus training epochs on object-related action with modifier (OA-M) recognition dataset. Others are the same as that in Fig. \ref{fig:comparison}.}
\label{fig:OA-M_error}
\end{center} 
\end{figure}

\begin{table}[t]
\caption{Accuracy comparison on OA-M recognition dataset}
\label{tab:exp2_accuracy}
\centering
\begin{tabular}{|c|ccc|c|}
\hline
\multirow{2}{*}{Model} & \multicolumn{4}{c|}{Accuracy}                                                               \\ \cline{2-5} 
                       & Object               & Action               & Modifier              &        Joint               \\ \hline
Spatial CNN & 87.9\% & 57.3\% & 36.7\% & 26.4\% \\ 
C3D & 97.8\% &	91.5 \% &	72.8\% &	68.8\%   \\ \hline\hline
Baseline     & 99.0 \% &	96.4 \% &	95.8 \% &	92.9\%   \\
AD     & 98.2\% &	98.4\% &	98.4\% &	95.4\%   \\
LN     & 97.6\% &	96.4\% &	95.0\% &	90.5\%   \\
LN+AD & 99.4\% &	98.0\% &	99.0\% &	97.2\%    \\ \hline
\end{tabular}
\end{table}

In this experiment, we do not use BN because of sequence length variability in a mini-batch. 
In Fig. \ref{fig:OA-M_error} and Table \ref{tab:exp2_accuracy}, we observe that LN is even worse than the baseline in terms of training speed and classification accuracy. We think that the statistics estimation error of LN is accumulated over time as we hypothesized in the previous experiment, and the error accumulation over longer periods of time leads to worse results for LN than that of the previous experiment. On the other hand, AD consistently offers faster convergence speed and higher classification accuracy. Furthermore, by solving the limitation of LN with the neuron-wise normalization of AD, LN+AD shows the most significant improvement of both training speed and generalization over the baseline, LN, and AD. 

The accuracy comparison of ConvGRU, spatial CNN, and C3D shows that temporal processing capability makes the clear distinction between the networks (see Table \ref{tab:exp2_accuracy}). Although the short-term processing capability of C3D is reasonable enough for object and action recognition, it is no longer true for modifier recognition that requires long-term information.
These results indicate that ConvGRU is essential for contextual video recognition. Hence, the proposed method, which can significantly accelerate the training of ConvGRU, becomes more crucial.

\section{Conclusion}
In this paper, inspired by detrending, we proposed a novel temporal normalization method, called \textit{adaptive detrending} (AD), to accelerate the training of recurrent neural networks (RNNs) by removing internal covariate shift over time. Although several normalization methods extended from batch normalization (BN) have been proposed for the training acceleration of RNNs, these methods utilize only the spatial domain not the time one, which is the target domain of RNNs, for normalization.
The key insight of this paper is considering the hidden state of gated recurrent unit (GRU) as a trend with an exponential moving average, so that we seamlessly implemented AD in GRU with a simple modification. AD has several benefits as follows: AD is highly efficient in terms of computational and memory requirements; a trend is automatically estimated through learning, unlike conventional detrending methods that are required manual parameter setting; and AD is generally applicable to both GRU and ConvGRU, which cannot be achieved by BN and layer normalization (LN).

In the experiments, we demonstrated that (1) convolutional GRU (ConvGRU) has a much richer temporal processing capability, which is crucial for contextual recognition, than the feed-forward neural networks, and (2) AD consistently provides faster convergence speed and better generalization than that of the baseline and spatial normalization methods. Also, we rediscovered that the bias initialization trick, specifically the negative bias initialization for the update gate of GRU, to address the slow and unstable learning problems of AD.
The qualitative analysis revealed that AD eliminates internal covariate shift, which explains the reason for the training acceleration; and AD controls the degree of detrending over time and samples, which explains the reason for the performance improvement.
Furthermore, AD gets an additional improvement by collaborating with the spatial normalization methods. Especially, a neuron-wise normalization by AD solves the main limitation of LN, which the assumption of LN is no longer true in the case of CNNs.

In conclusion, AD substantially alleviates the computational burden of ConvGRU by accelerating the training speed with little additional cost, and shows strong synergy with the existing normalization methods.
Therefore, AD would be helpful in future studies that want to utilize a rich spatio-temporal processing capability of ConvGRU and its variants.
For future work, we will apply AD to speech recognition. The step-wise BN and LN \cite{Cooijmans:2016,Ba:2016} are hard to apply for speech recognition because these might lose the dynamics of speech signals. The sequence-wise BN \cite{Laurent:2015} has demonstrated the training acceleration and performance improvement in speech recognition \cite{Amodei:2016} by preserving the speech signals' dynamics, but it cannot provide different degrees of normalization over time, which might be required for further eliminating internal covariate shift. An automatic control for the degree of detrending by AD is expected to find the balance between the preservation for the dynamics of speech signals and the reduction of internal covariate shift.

\appendices
\section{Convolutional Neural Network}
\label{appx:CNN}

\begin{table}[t]
\centering
\caption{Convolutional neural network configuration. The format of the table is the same as that in Table \ref{tab:architecture}}
\label{tab:CNN}
\begin{tabular}{|c|l|l|c|c|}
\hline
\multirow{2}{*}{Layer} & \multicolumn{1}{c|}{\multirow{2}{*}{Type}} & \multicolumn{1}{c|}{\multirow{2}{*}{Filter}} & \multirow{2}{*}{Stride} & \multirow{2}{*}{Pad} \\
                       & \multicolumn{1}{c|}{}                      & \multicolumn{1}{c|}{}                        &                         &                      \\ \hline\hline
1                      & Conv (ReLU)                                & 7$\times$7$\times$3$\times$96                & 2$\times$2                       & 0$\times$0                    \\ \hline
2                      & Max                                        & 2$\times$2                                   & 2$\times$2                       & 0$\times$0                    \\ \hline
3                      & Conv (ReLU)                                & 5$\times$5$\times$96$\times$256              & 2$\times$2                       & 1$\times$1                    \\ \hline
4                      & Max                                        & 2$\times$2                                   & 2$\times$2                       & 0$\times$0                    \\ \hline
5                      & Conv (ReLU)                                & 3$\times$3$\times$256$\times$512             & 1$\times$1                       & 1$\times$1                    \\ \hline
6                      & Conv (ReLU)                                & 3$\times$3$\times$512$\times$512             & 1$\times$1                       & 1$\times$1                    \\ \hline
7                      & Conv (ReLU)                                & 3$\times$3$\times$512$\times$512             & 1$\times$1                       & 1$\times$1                    \\ \hline
8                      & Global Avg                                 & 6$\times$6                                   & -                       & -                    \\ \hline
\multirow{3}{*}{9}     & FC (Softmax)                               & 1$\times$1$\times$512$\times$$C_{1}$               & -                       & -                    \\ \cline{2-5} 
                       & \multicolumn{4}{c|}{\setstackgap{S}{.7pt}\Shortstack{. . .}}                                                                                                                     \\ \cline{2-5} 
                       & FC (Softmax)                                           & 1$\times$1$\times$512$\times$$C_{N}$               & -                       & -                    \\ \hline
\end{tabular}
\end{table}

We follow the architecture used in the spatial CNN \cite{Simonyan:2014}. However, we replace the last max pooling and two fully connected layers with a global average pooling layer. The networks consist of five convolutional (Conv), two max pooling (Max), one global average pooling (Global Avg), and one fully-connected (FC) layers. Details of the network configuration are shown in Table \ref{tab:CNN}.

Following Section \ref{sec:training} for training, stochastic gradient descent (SGD) with Nesterov momentum 0.9 \cite{Nesterov:1983,Nesterov:2004} is applied to train a network; the size of mini-batch, L2-norm weight decay, and threshold of gradient clipping is set to 8, 0.0005, and 10, respectively.
All weights and biases of a network are initialized from a zero-mean Gaussian distribution with a standard deviation of 0.03. 
Also, data pre-processing and augmentation methods are exactly the same as Section \ref{sec:preprocessing}.
All networks are trained with a learning rate of 0.01 over 200 epochs on object-related action dataset and over 300 epochs on object-related action with modifier dataset.

We follow the evaluation protocol used in the spatial CNN \cite{Simonyan:2014}.
In detail, 25 frames are sampled with equal spacing from a video, and 10 crops (1 center, 4 corners, and their horizontal flipping) are obtained from each sampled frame. Then, the scores are averaged across the sampled frames and crops of each frame to obtain the final classification accuracy for a video.

\section{Convolutional 3D}
\label{appx:C3D}

\begin{table}[t]
\centering
\caption{Convolutional 3D configuration. The format of the table is the same as that in Table \ref{tab:architecture} except that time is added as the first dimension of the filter, stride, and pad}
\label{tab:C3D}
\begin{tabular}{|c|l|l|c|c|}
\hline
\multirow{2}{*}{Layer} & \multicolumn{1}{c|}{\multirow{2}{*}{Type}} & \multicolumn{1}{c|}{\multirow{2}{*}{Filter}}  & \multirow{2}{*}{Stride}                  & \multirow{2}{*}{Pad}                     \\
                       & \multicolumn{1}{c|}{}                      & \multicolumn{1}{c|}{}                         &                                          &                                          \\ \hline\hline
1                      & 3D Conv (ReLU)                             & 3$\times$7$\times$7$\times$3$\times$32        & 1$\times$2$\times$2                      & 1$\times$0$\times$0                      \\ \hline
2                      & 3D Max                                     & 2$\times$2$\times$2                           & 2$\times$2$\times$2                      & 0$\times$0$\times$0                      \\ \hline
3                      & 3D Conv (ReLU)                             & 3$\times$5$\times$5$\times$32$\times$64       & 1$\times$2$\times$2                      & 1$\times$1$\times$1                      \\ \hline
4                      & 3D Max                                     & 2$\times$2$\times$2                           & 2$\times$2$\times$2                      & 0$\times$0$\times$0                      \\ \hline
5                      & 3D Conv (ReLU)                             & 3$\times$3$\times$3$\times$64$\times$128      & 1$\times$1$\times$1                      & 1$\times$1$\times$1                      \\ \hline
6                      & 3D Max                                     & 2$\times$2$\times$2                           & \multicolumn{1}{l|}{2$\times$2$\times$2} & \multicolumn{1}{l|}{0$\times$0$\times$0} \\ \hline
7                      & 3D Conv (ReLU)                             & 3$\times$3$\times$3$\times$128$\times$256     & \multicolumn{1}{l|}{1$\times$1$\times$1} & \multicolumn{1}{l|}{1$\times$1$\times$1} \\ \hline
8                      & 3D Global Avg                              & 2$\times$3$\times$3                           & -                                        & -                                        \\ \hline
\multirow{3}{*}{9}     & FC (Softmax)                               & 1$\times$1$\times$1$\times$256$\times$$C_{1}$ & -                                        & -                                        \\ \cline{2-5} 
                       & \multicolumn{4}{c|}{\setstackgap{S}{.7pt}\Shortstack{. . .}}                                                                                                                                                            \\ \cline{2-5} 
                       & FC (Softmax)                               & 1$\times$1$\times$1$\times$256$\times$$C_{N}$ & -                                        & -                                        \\ \hline
\end{tabular}
\end{table}

Convolutional 3D (C3D) \cite{Tran:2015} consists of four 3D convolution (3D Conv), three 3D max pooling (3D Max), one 3D global average pooling (3D Global Avg), and one fully-connected (FC) layers. Because C3D receives $L$ consecutive frames as an input to process short-term information, convolution and pooling operations are extended from 2D (spatial) to 3D (spatio-temporal). Details of the network configuration are shown in Table \ref{tab:C3D}.

Details of (1) training and (2) data pre-processing and augmentation are the same as Appendix \ref{appx:CNN} except that the number of stacked frames $L$ set to 16.
All networks are trained with a learning rate of 0.02 over 200 epochs on object-related action dataset and with a learning rate of 0.01 over 300 epochs on object-related action with modifier dataset. 

We follow the evaluation protocol used in the temporal CNN \cite{Simonyan:2014}. In detail, we sample equally spaced five video sub-volumes, each has 16 consecutive frames. For each selected sub-volumes, 10 crops are obtained by cropping 1 center and 4 corners, and horizontally flipping them. The final classification accuracy for a video is obtained by averaging the scores across the sampled sub-volumes and crops of each sub-volumes.


\ifCLASSOPTIONcaptionsoff
  \newpage
\fi



\bibliographystyle{IEEEtran}
\bibliography{ref}
%



%

\begin{IEEEbiography}[{\includegraphics[width=1in,height=1.25in,clip,keepaspectratio]{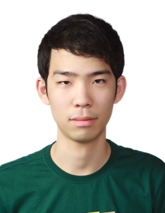}}]{Minju Jung}
received the B.E. degree in electronic engineering from Hanyang University, Seoul, Republic of Korea, in 2013. 
He is currently pursuing the Ph.D. degree in electrical engineering Department with Korea Advanced Institute of Science and Technology, Daejeon, Republic of Korea.

His current research interests include recurrent neural networks and computer vision.
\end{IEEEbiography}

\begin{IEEEbiography}[{\includegraphics[width=1in,height=1.25in,clip,keepaspectratio]{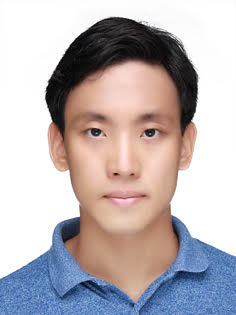}}]{Haanvid Lee}
received the B.S. degree in electrical and electronic engineering from Yonsei University, Seoul, Republic of Korea, in 2015, and the M.S. degree in electrical engineering from Korea Advanced Institute of Science and Technology, Daejeon, Republic of Korea, in 2017.

His current research interests include computer vision, recurrent neural networks, and reinforcement learning.
\end{IEEEbiography}


\begin{IEEEbiography}[{\includegraphics[width=1in,height=1.25in,clip,keepaspectratio]{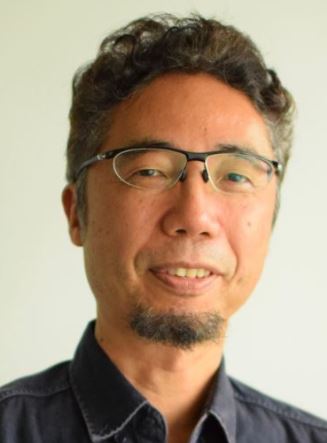}}]{Jun Tani}
received a doctor of engineering degree in electrical engineering from Sophia University, Tokyo, Japan, in 1995. 

He worked for Sony Corporation, Tokyo, Japan, and later for Sony Computer Science Lab, Tokyo, Japan, as a researcher from 1990 to 2001. Then, he worked at Riken Brain Science Institute, Saitama, Japan, from 2001 to 2012, where he has been a PI of Lab. for Behavior and Dynamic Cognition. He had been also appointed as a Visiting Associate Professor at University of Tokyo, Tokyo, Japan, from 1997 to 2002. He became a full professor of Electrical Engineering in Korea Advanced Institute of Science and Technology, Daejeon, Republic of Korea, in 2012, where he started Cognitive Neuro-Robotics Lab. He has been appointed also as a visiting professor at Waseda University, Tokyo, Japan, and an adjunct professor at Okinawa Institute of Science and Technology, Okinawa, Japan, since 2014 and 2017, respectively. 

His research interests include neurorobotics, deep learning, complex systems, brain science, developmental psychology, and philosophy of mind. He is an author of ``Exploring Robotic Minds: Actions, Symbols, and Consciousness as Self-Organizing Dynamic Phenomena." published from Oxford University Press.
\end{IEEEbiography}




\end{document}